\documentclass{iosart2x}

\usepackage{dcolumn}
\usepackage{dsfont}
\usepackage{mathtools}
\usepackage[linesnumbered, ruled, vlined]{algorithm2e}
\usepackage{tikz}

\newcolumntype{d}[1]{D{.}{.}{#1}}

\firstpage{1}
\lastpage{5}
\volume{1}
\pubyear{2018}

\begin{document}
\begin{frontmatter} 

%
  \title{Automatic Processing and Solar Cell Detection in Photovoltaic
    Electroluminescence Images
  }

\runningtitle{Automatic EL processing}

\author{\inits{E.}\fnms{Evgenii} \snm{Sovetkin}\ead[label=e1]{sovetkin@stochastik.rwth-aachen.de}%
}%
 \ and %
\author{\inits{A.}\fnms{Ansgar} \snm{Steland}\ead[label=e2]{steland@stochastik.rwth-aachen.de}}\thanks{Corresponding author. \printead{e2}.}
\runningauthor{E. Sovetkin et al.}
\address{Institute of Statistic, \orgname{RWTH Aachen University}, W\"ullnerstr.\ 3, D-52062, Aachen, \cny{Germany}\printead[presep={\\}]{e1}}
\address{\printead{e2}}

\begin{abstract}
  Electroluminescence (EL) imaging is a powerful and established
  technique for assessing the quality of photovoltaic (PV) modules,
  which consist of many electrically connected solar cells arranged in
  a grid. The analysis of imperfect real-world images requires
  reliable methods for preprocessing, detection and extraction of the
  cells.  We propose several methods for those tasks, which, however,
  can be modified to related imaging problems where similar geometric
  objects need to be detected accurately. Allowing for images taken
  under difficult outdoor conditions, we present methods to correct
  for rotation and perspective distortions.  The next important step
  is the extraction of the solar cells of a PV module, for instance to
  pass them to a procedure to detect and analyze defects on their
  surface. We propose a method based on specialized Hough transforms,
  which allows to extract the cells even when the module is surrounded
  by disturbing background and a fast method based on cumulated sums
  (CUSUM) change detection to extract the cell area of single-cell
  mini-module, where the correction of perspective distortion is
  implicitly done. The methods are highly automatized to allow for big
  data analyses. Their application to a large database of EL images
  substantiates that the methods work reliably on a large scale for
  real-world images. Simulations show that the approach achieves high
  accuracy, reliability and robustness.  This even holds for low
  contrast images as evaluated by comparing the simulated accuracy
  for a low and a high contrast image.
\end{abstract}

\end{frontmatter}

\section{Introduction}

The automatic preprocessing of industrial images of objects is an
ubiquitous and important task of quality control, in order to prepare
object extraction, identification and its assessment as well as
comparisons with reference images. For example, this applies to
high-resolution imaging technologies to check materials on a
microscopic or nano scale, where position and orientation of the
sensor and the sample cannot be perfectly controlled. The same issue
arises when the object of interest is surrounded by background such as
a fluid or gas, or when images are taken in difficult environments,
for example under outdoor (or field) conditions. The latter situation
motivated and initiated the procedures proposed in this paper.

The preprocessing methods proposed and investigated in this article
are tailored to electroluminescence (EL) images of photovoltaic
modules and cells, especially when taken under outdoor conditions, but
they are applicable to various similar imaging problems, such as the
analysis of wafers in CPU production, with appropriate
modifications. Nevertheless, in our exposition we will focus on the
application to imaging in photovoltaics. The approach presented in
this paper has been used to preprocess a large number of EL images
collected in a panel-type multi-site study \cite{FischerEtAl2018}, in
order to prepare them for further statistical analyses. Tailored
multiple regression methods have been proposed by \cite{PepSteSov2017} and
applied to images preprocessed using the method proposed here. In practice, one often computes
certain quality features from the sampled images, and it is even possible to calculate the operating voltage of individual solar
cells by EL imaging, \cite{PotthoffEtAl2010}. When focusing
on the problem to accept or reject a lot of photovoltaic modules (PV modules)  based on statistics derived from such images, we refer to \cite{AlthausHerrmannStelandZaehle06, HerrmannSteland2010, StelandZaehle2009, AvellanPepelyshevSteland2013}. Extensions to two-point inspection schemes have been proposed by
 \cite{Steland2015a}.

In the photovoltaic industry, imaging is a widely established tool
to assess and inspect the quality of PV modules and solar
cells. For a general overview and references to established methods aiming at detecting certain
defects and issues such as macro-crack detection using anisotropic diffusion as in machine vision,
\cite{TsaiEtAl2010} or inspection of electrical contacts, \cite{SunEtAl2010}, we refer to \cite{Mauk}. 
EL is one of the most established imaging technologies. It allows a microscopic view into the crystalline cell material and therefore allows to detect faults such as micro cracks, cell breaks or
interconnect and soldering flaws, which are usually invisible
otherwise, since they are only in effect when the photovoltaic effect
is active. EL inverts that photovoltaic effect and allows to image the
spatial distribution minority carrier diffusion length,
\cite{Fuyuki2005}: If a solar cell is supplied by a DC current,
radiative recombination occurs and emits photons (i.e. luminescence),
which can be captured by a charge coupled device (CCD) camera that is
sensitive to the relevant spectrum. Here, long exposure times are
required, such that even infinitesimal movements of the camera or the
PV module, for example due to wind, vibrancy or shocks, may lead to
reduce image sharpness and quality.

Despite the progress, the evaluation of an EL image is usually based on expert
knowledge only, \cite{EvansSugiantoMao2014}. This is mainly due to the lack of appropriate computerized,
automatic methods for image processing and advanced image analysis, which requires accurate detection of the
relevant cell areas in an image, in order to allow sound statistical analyses of defect. Otherwise, even simple statistical
measures such as the percentage of low performing cell area can not be determined, and difference images will show artifacts leading to false detections. These issues, which are of particular relevance for outdoor imaging, call for tailor-made preprocessing methods as proposed in this paper.

When images are collected under field conditions, the required outdoor
acquisition procedure leads to several problems that are usually not
present in a lab environment. In brief, the process is as follows:
First, each PV module is disconnected from the grid. Then, either it
is unmounted and carried to a mobile lab or a frame is mounted, which
holds the camera and protective curtains to shield the module from
external light sources and allow for long-exposure imaging.  That
procedure is highly sensitive to mechanical deficiencies and stress
(incorrect mounting of the frame, imperfect positioning of the camera,
vibrancy due to wind, shocks, residual external light etc.) resulting
in several image processing problems. Firstly, the module is
incorrectly shown on the image, as it can be rotated in all three
dimensions, such that it is neither parallel to the camera's sensor
nor its center coincides with the center of the sensor. Secondly,
unlike images taken in a laboratory where a fixed and calibrated
environment is used, the module's position differs from image to
image. See examples of typical EL images in
Section~\ref{sec:simulations-1}.

In this paper, we discuss novel methods that allow to correct for
rotation and perspective distortions, to estimate the boundaries of
the PV module and to extract all PV cells, which are the relevant
areas where the photovoltaic effect occurs. To the best of our
knowledge, there is no published method for those problems. One could
try to apply methods from machine vision and learning,
\cite{Davies2004, Goodfellow2016}, especially deep learning
neural networks which allow to approximate square integrable
\cite{ShahamEtA2018}. For image processing and analysis, convolutional layers
are used to extract local information which is processed in later layers to detect geometrical shapes. 
The trained deep learner is then used as a surrogate model for the unknown optimal classification
function and approximates the latter under weak conditions, see
\cite{Steland2018}. Recent studies, e.g. \cite{KC2017}, provided evidence that image classification by
such deep learners can be improved by data augmentation or denoising prior to classification. 
But such approaches, especially data augmentation to deal with noise (see \cite[Sec.~6]{KC2017}), require a large learning
sample with known - manually determined - correct
positions of the solar cells. For the problem at hand this is not feasible.  
One could also draw on methods developed for the detection of
quadrilateral documents in images, see \cite{Fan2016} and the
references therein. The detection of solar cells in EL images
differs in that the number of quadrilateral areas is known
and fixed, whereas in document processing either one quadrilateral is
determined or, when aiming at the detection of paragraphs etc., it is
estimated. Hence the number of detected areas and their relative
position may severely depend on the quality of the image. Contrary,
the problem studied here is different and our approach makes use of
the specific structure and always extracts the right number of areas.

In our approach, we rely on two basic approaches. Firstly, we use the Hough transform, 
a general approach to locate objects in images, see
the review \cite{Davies2008}. We adopt and specialize it to the
specific case of EL images of PV modules and combine it with statistical robust
regression as well as a physical knowledge about optical distortions,
namely rotation and perspective distortion. In practice, a further
issue arises, namely radial distortion caused by the camera's
lens. There are fairly standard methods to correct for it. For details
we refer to \cite{Hartley2007parameter}. In practice, it is, however,
not always possible to correct for this distortion, and therefore we
propose preprocessing methods that are to some extent robust with
respect to this issue. The second approach applies the CUSUM
change-point estimator, see \cite{Hawkins1977}, to detect the
boundaries of a solar cell.  The CUSUM estimator is the Likelihood
estimator for independent Gaussian data assuming common means before
and after the change. It can also be interpreted as optimally fitting
a model where it is assumed that there are exactly two different line
segments (foreground = cell area and background), which are separated
by the change-points. To the best of our knowledge, the application of
that approach to the problem of detecting solar cells in PV modules
has not been studied yet.

We discuss two preprocessing work-flows. The first one applies the
tasks of rotation correction, perspective correction and cell
extraction in a sequential way. It is especially suited for PV modules,
which consist of many solar cells. The second approach is a simplified
and fast procedure, mainly designed for one-cell modules, which
extracts the cell area from the raw image and then applies an
appropriate transformation to output a rectangular shaped image of the
cell area.

The paper is organized as follows. Section 2 reviews the Hough
transform often used to detect lines or, more generally, parametric
curves. In Section 3, we describe a method that allows to correct for
the rotation of a PV module. Section 4 discusses a method for the
correction of perspective distortion. In Section 5, we consider a
specialized Hough transform and propose a method for identifying the
location of a PV module and its cells, such that they can be extracted
from the image. The simplified fast procedure for one-cell modules is
presented in Section 6. Lastly, Section 7 applies the methods and
algorithms to real images.

\section{Hough Transform}
\label{sec:hough-transform}

The Hough transform is a technique to detect objects such as lines or
circles in images, see \cite{Davies2008}.  It is widely known and used
in computer vision applications, \cite{illingworth1988survey,
  2004Goldenshluger}. For its efficient computation, several
algorithms have been developed, see \cite{matas2000robust}. It has
also been extended to identify parametric curves as discussed and
applied in \cite{ballard1981generalizing}.

As a preparation of the proposed algorithms detailed in the subsequent
sections and to introduce required notation, we give a brief review of
a generic Hough transform. The Hough transform method is applied to a
binary image, for example obtained by thresholding a gray scale
image. We consider a binary image as a subset
$\Omega \subset \mathcal{D}$, where $\mathcal{D}$ is a finite subset
of $\mathbb{N}^{2}$ and $\mathbb{N}$ denote the natural numbers. Each
element $\omega\in\Omega$ corresponds to the coordinates of a non-zero
pixel of binary image.

We call $\mathcal{H}$ a {\em Hough space} or {\em Hough domain}, if
its elements are parameters representing a curve of interest. For
example, if $h = (a,b) \in \mathcal{H}$ is a two-dimensional vector,
it represents a line, $y = ax + b$, with intercept $b$ and slope $a$
in the image $\Omega$.

The Hough Transform is defined as a mapping which assigns to each
coordinate $x \in \mathcal{D}$ a set $H_{x}$ of curves that go through
the point, that is a subset of $\mathcal{H}$,
\begin{align*}
  \mathcal{P}: \mathcal{D} \to 2^{\mathcal{H}}, \mathcal{P}(x) = H_{x}.
\end{align*}
where $2^{\mathcal{H}}$ is the powerset of $\mathcal{H}$.  In case of
the Hough line transform, we assign a set of lines going through the
point $x$.

For a given image we are interested in identifying a single or several
parameters of the Hough space, namely those which should be regarded
as curves (or lines) present in the image. The identification of those
elements in the Hough space is achieved by solving the following
optimization problem.

Let $R$ be a positive number and $B_{R}(h)$ be a neighbourhood of a
point $h$ in the space $\mathcal{H}$. Then, the optimization problem
is given by
\begin{gather}
  \label{eq: hough optimisation}
  \sum\limits_{x \in \Omega}
  \mathds{1}
  \left\{
    B_{R}(h) \cap \mathcal{P}(x) \neq \emptyset
  \right\}
  \to \max\limits_{h \in \mathcal{H}}.
\end{gather}
Here $ \mathds{1}\{A\} $ is equal to $1$, if the expression $A$ is true,
and $0$ otherwise.

The neighbourhood $B_{R}(h)$ is usually selected to be an Euclidean
ball with radius $R$. Its role is to control the accuracy of the
method in finding a curve. Points of the image which are in the
neighbourhood are still regarded as lying on the parametric curve
given by $h$.  One is interested in a global optimum of \eqref{eq:
  hough optimisation} when searching for a unique curve. Finding
multiple local maxima corresponds to the detection of multiple curves.

The optimization problem \eqref{eq: hough optimisation} is solved by
using Algorithm~\ref{algo: hough optimistion}.

\begin{algorithm}
  \DontPrintSemicolon Select a grid of point in the space
  $\mathcal{H}$. Associate to each grid point a counter;\\
  For each non-zero pixel location $x \in \Omega$ of the binary image
  evaluate the Hough transform $H_{x} = \mathcal{P}(x)$;\\
  For each coordinate $h \in H_{x}$ increment its associated counter;\\
  The result is evaluated by identifying the grid points with the
  maximal counter value and those where local maxima are present.
  \caption{Hough optimisation algorithm}
  \label{algo: hough optimistion}
\end{algorithm}

\section{Rotation distortion correction}
\label{sec:rotat-dist-corr}

The procedure to correct for rotation of a PV module in an EL image is
pursued by definining an objective function that attains its maximum
when the PV module is correctly positioned in the image. That
objective function uses the fact that after the optimal rotation
the horizontal and vertical lines present in the PV module induce a
pattern in the row and column sums.

Formally, let $I$ be a gray scale image, i.e.\ a function
$I: \{1,\ldots,W\}\times\{1,\ldots,H\} \to \mathbb{R}$, where $H$ is
the height and $W$ is the width of the image $I$. Define the following
two vectors $R^{I} \in \mathbb{R}^{H}$, $C^{I} \in \mathbb{R}^{W}$:
\begin{gather*}
  R^{I}_{j} = \sum\limits_{i=1}^{W} I(i,j), \quad \text{for } j\in\{1,\ldots,H\},\\
  C^{I}_{i} = \sum\limits_{j=1}^{H} I(i,j), \quad \text{for } i\in\{1,\ldots,W\}.
\end{gather*}
$ R^{I} $ are the column sums of the pixel values and $ C^{I} $ are the row
sums.

A PV module is correctly positioned in an EL image when all grid lines
of the PV module are parallel to the edges of the image. In that case,
since the grid lines are darker than the active crystalline area, we
would then observe negative peaks in the vectors $R^{I}$ and
$C^{I}$. However, when a PV module is rotated then the pixels
corresponding to each grid line appear in different coordinates of the
vectors $R^{I}$ and $C_{I}$, and hence, no negative peaks are present
in $R_{I}$ and $C_{I}$. Therefore, identifying the optimal rotation
angle (i.e. when vectors $R^{I}$ and $C^{I}$ contain largest peaks)
can be done by maximising variances of the vectors $R^{I}$ and
$C^{I}$.

Define $T_{\alpha}(I)$ to be an operator that rotates the image $I$ by
the angle $\alpha$ around the image centre. Then, the objective
function $f$ is defined by
\begin{gather*}
  f(T_{\alpha}(I)) \coloneqq \mathrm{sd}(R^{T_{\alpha}(I)}) +
  \mathrm{sd}(C^{T_{\alpha}(I)}),
\end{gather*}
where $\mathrm{sd}$ is the sample standard deviation of a vector,
i.e.,\ for $x = (x_1, \dots, x_n)' \in\mathbb{R}^{n}$
\begin{gather*}
  \mathrm{sd}(x) =
  \sqrt{
    \frac{1}{n}
    \sum\limits_{i=1}^{n}
    (x_{i} - \bar x)^{2}
  },
\end{gather*}
and $\bar x$ is the sample mean. In order to correct for the PV module
rotation, we solve the following optimization problem
\begin{gather}
  \label{eq:1}
  f(T_{\alpha}(I)) \to \max_{\alpha \in [-\pi/4,\pi/4]}.
\end{gather}

The optimization problem (\ref{eq:1}) is solved with a simple golden
section optimisation algorithm (see \cite{kiefer1953sequential}), as
it does not require any additional assumptions on an image $I$ or the
function $f$.  The golden section algorithm is given in
Algorithm~\ref{algo: the golden-section}.

\begin{algorithm}
  Let $f$ be an objective function to be maximised on an interval
  $[a,b]$. Evaluate $f$ at three probe points $a$, $b$ and $x_{0}$,
  where $x_{0}$ is the smallest of the two points satisfying the
  golden-ratio relation
  $\frac{b-a}{b-x_{0}} = \frac{b-x_{0}}{x_{0}-a}$. \\
  Evaluate $f$ at the second golden-ratio point $x_{1}$. \\
  By comparing four evaluated function values select either the
  interval $[a,x_{1}]$
  or $[x_{0},b]$ to be a new search interval. \\
  Repeat Step 2 until the interval length is smaller than a selected
  $\varepsilon$. The golden-ratio choice of the probe points
  guarantees that the three probe points in the new interval also
  satisfies the golden ratio relation. Therefore, at each step only a
  single additional function evaluation is performed.
  \caption{The golden-section optimisation algorithm}
  \label{algo: the golden-section}
\end{algorithm}

Observe that the algorithm does not require a preprocessing step to
detect the edges, which would increase the computational burden. The
algorithm can be applied directly to an EL image.

To summarise, the steps of the PV module rotation correction algorithm are
given in Algorithm~\ref{algo: rotation}.

\begin{algorithm}
  \DontPrintSemicolon
  Solve optimization problem (\ref{eq:1}) using the golden section
  algorithm (Algorithm~\ref{algo: the golden-section}) and obtain the
  optimal rotation angle $\alpha_{0}$;\\
  Rotate the image by the angle $\alpha_{0}$.
  \caption{PV module rotation correction}
  \label{algo: rotation}
\end{algorithm}

\section{Perspective distortion correction}
\label{sec:persp-dist-corr}

The module rotation transformation discussed in the previous section
models a 2-dimensional rotation of a PV module in an EL image; it
rotates the optical image of the module on the camera's sensor. However, a
PV module can be rotated in all three dimensions which results in a
so-called perspective distortion. Physically that distortion occurs
when a PV module surface is not located perpendicular to the focus
line of a camera. In this section, we present a method to correct for
this type of distortion. As a result, the image is transformed in such
a way that all PV module grid lines are parallel to the edges of an EL
image.

Correction for perspective distortion is a common subject in image
processing  (see \cite{frazier1997automated}). In most cases, those
algorithms employ easily detectable markers on an image and use
their coordinates to estimate the perspective distortion
parameters. In our application, placing some markers on PV module is not
feasible, and therefore, we need to develop an alternative approach.

The perspective distortion is physically modelled by 8 parameters (see
\cite{tan1993computing}). It preserves straight lines.  Therefore, the  grid lines  of the PV module appear
as straight (though not necessarily parallel) lines in an EL image.

In order to correct for perspective distortion, we could follow the
same strategy as we use for the correction of a module rotation, see
Section~\ref{sec:rotat-dist-corr}. However, the resulting optimization
problem is more complex as it contains $8$ parameters. Numerical
studies indicated that the formulated objective function often has
local minima that do not satisfy our objective,
which impede its application in practice.

Therefore, our strategy of correction for perspective distortion is as
follows. We estimate grid lines using the Hough line
transform and use them to estimate the perspective distortion
parameters.  Note that it suffices to known two vertical and two
horizontal lines to determine the perspective parameters. Our
estimator, however, incorporates all detected grid lines.  In this
way, the influence of incorrectly detected lines is reduced and the
method is robust against this type of disturbance.

In order to apply the Hough line transform, we need to obtain a binary
image. Instead of simple thresholding of a given image, we use
specific versions of the USAN edge detection algorithm,
\cite{2007Rafajlowicz, RPS2008, Steland2015} in order to improve the
accuracy of the detection. The USAN approach aims at reproducing edges
and simultaneously smoothes homogeneous areas, which increases the
signal-to-noise ratio and leads to improved line detection. A further
important motivation for choosing the USAN edge detector is that it
can be designed to output thick edges. This is beneficial when the
grid lines do not appear as straight lines, e.g.\ due to the radial
distortion effect or low contrast. In this way the proposed method is
robust with respect to radial distortion.

We apply the USAN edge detection twice with two different
kernels. Firstly, we apply the edge detector with vertically
positioned rectangular kernel, so that its sensitivity with respect to
vertically positioned edges is high, whereas its sensitivity to detect
horizontal lines is low. Secondly, we apply the USAN edge detector
with horizontally positioned rectangular kernel, which is tuned to
detect horizontally positioned edges. As a result, we obtain two
binary images: one image with horizontal and another one with vertical edges.

The Hough line transform is then applied to the resulting edge
images. The binary image with horizontal edges is used to detect the
horizontal grid lines, whereas the second image with vertical edges
yields the vertical grid lines. Recall that the result of the Hough
line transform is a collection of lines. In order to avoid non-grid
line detections, we remove those vertical and horizontal lines which
have slopes and angles, respectively, lying outside preselected
intervals.

At this stage of the procedure we obtain two data sets consisting of
the detected horizontal and vertical lines.  For each of those data
sets linear regressions are conducted, where the independent variable
is the ordinate of a vertical lines (or an abscissa in case of
horizontal lines) and the dependent variable is the corresponding line
slope angle. For details and illustrations we refer to
\cite{SovSte2015}.  As a result, we obtain estimates of the angle of a
vertical (horizontal) grid line's slope as a function of $x$ ($y$).

It remains to estimate the perspective distortion parameters from
those functions. Obviously, it is sufficient to specify the
coordinates of two quadrangles (see Figure~\ref{fig: correcting
  rectangular}). For that construction fix 4 points $M$, $N$, $P$ and
$Q$. The first quadrangle is the square $A'B'C'D'$. The second
quadrangle is obtained by considering the intersection of the
following four lines. The first line is the line passing through the
point $M$ with a slope equal to the value predicted by the simple
linear regression for vertical lines, evaluated at the ordinate of the
point $M$. The second line is the line passing through the point $N$
with a slope equal to the predicted value resulting from the simple
linear regression for horizontal lines, evaluated at the abscissa of
the point $N$. The remaining lines corresponding to the points $P$ and
$Q$ are calculated analogously.

The resulting coordinates of the two rectangles are now used to feed
the perspective distortion procedures. In our implementation we use
the corresponding function from the ImageMagick
library~\cite{imagemagick}.

\begin{figure}[htb] \centering
  \begin{tikzpicture}[x=14mm,y=14mm]
    \draw[very thin] (-1,-1) node [below left] {A'}
    -- (-1,1) node [above left] {B'}
    -- (1,1) node [above right] {C'}
    -- (1,-1) node [above left] {D'}
    -- cycle;
    \draw (-1.09,-0.9) node [left] {A}
    -- (-0.905,0.95) node [below right] {B}
    -- (0.84, 1.046409) node [above left] {C}
    -- (1.1693067,-1.1072753415) node [right] {D}
    -- cycle;
    \fill (0,0) circle [radius=0.02] node [below left] {O};
    \draw [dashed] (-1,0) node [below left] {M}
    -- (1,0) node [below right] {P};
    \draw [dashed] (0,1) node [above left] {N}
    -- (0,-1) node [below left] {Q};
  \end{tikzpicture}
  \caption{Scheme of correcting the rectangular}
  \label{fig: correcting rectangular}
\end{figure}
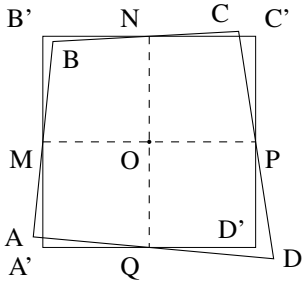

The method of perspective distortion correction discussed here
combines USAN edge detection and Hough line transform
sub-procedures. These procedures have several tuning parameters. The
question arises as to how one should select those parameters.  In
order to be able to use one set of tuning parameters for every EL
image, we standardize the image (subtract mean and divide by standard
deviation of pixel value intensities) and select USAN edge detector
and Hough line transform parameters manually by trial-and-error.

We summarise the perspective distortion correction procedure in
Algorithm~\ref{algo: perspective distortion}.

\begin{algorithm}
  \DontPrintSemicolon
  Standardise image; \\
  Apply vertical and horizontal USAN edge detection yielding $I_{v}$ and $I_{h}$; \\
  Apply Hough line transform to the binary images $I_{v}$ and $I_{h}$; \\
  Threshold vertical and horizontal lines with respect to the allowed
  slope angle intervals; \\
  Compute two simple linear regressions for the vertical and horizontal slopes; \\
  Use these linear regressions to compute the coordinates of quadrangles vertices  in Figure~\ref{fig: correcting rectangular}; \\
  Apply perspective distortion transformation using the computed
  quadrangles vertices.
  \caption{PV module perspective distortion correction}
  \label{algo: perspective distortion}
\end{algorithm}


\section{Cell detection}
\label{sec:grid-line-detection}

In this section we describe a specialized version of the Hough
transform discussed in Section~\ref{sec:hough-transform}, which aims
at identifying the location of a PV-module and all solar cells in an
EL image. Whereas the Hough transform is often used to detect small,
localized objects of given shape like circles, \cite{Davies2008}, here
we develop a Hough transform to identify a complex structure (the
module and its grid line structure separating the cells) stretched out
over a large part of the image. We use the Hough transform to seek
that pattern.

Recall from our discussion above that parallel lines  may be
bent in a real EL image, due to the effect of radial
distortion. Further, a PV module is usually rotated and not positioned
perpendicular to the camera's focus axis. The latter results in
rotation and perspective distortions. Although it could be possible to
design an algorithm based on the Hough transform that takes into
account the radial, the rotation and the perspective distortion
parameters simultaneously, the resulting model contains too many
parameters and the Hough transform optimization problem becomes hard
to compute.

In the first pre-processing step, we first apply the radial distortion
correction procedure as described in \cite{Hartley2007parameter}, and
the rotation and perspective distortion correction procedures
discussed in the previous sections. Thus, the problem to identify the
positions of the module and its cells in the image can be reduced to
finding the pattern of grid lines, i.e.  a fixed number of vertical
and horizontal lines in the EL image. The problem can be split into
two problems by separately detecting the vertical and horizontal
lines.

\begin{figure}[htb]
  \centering
  \begin{tikzpicture}[x=5mm,y=3mm]
    \draw (0, 0) -- (0, 8);
    \draw (2, 0) -- (2, 8);
    \draw (4.5, 0) -- (4.5, 8);
    \draw (7.5, 0) -- (7.5, 8);
    \draw[<->] (0,4) -- (2,4);
    \draw[<->] (2,4) -- (4.5,4);
    \draw[<->] (4.5,4) -- (7.5,4);
    \node[above] at (1,4) {$\Delta_{0}$};
    \node[above] at (3.25,4) {$\Delta_{1}$};
    \node[above] at (6,4) {$\Delta_{2}$};
  \end{tikzpicture}
  \caption{The (relative) distances between the grid lines of a
    PV-module are known. This information can be used for their
    detection in an EL image.}
  \label{fig: several parallel lines}
\end{figure}
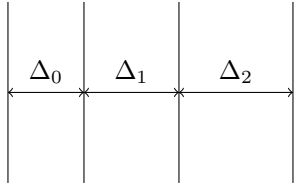

Since the physical dimensions of a PV module are known, and so are the
distances between the grid lines relative to the size of module, the
problem of cell detection can be formulated as follows: detect $n$
parallel lines (see Figure~\ref{fig: several parallel lines})
separated by intervals that are equal to
$\Delta_{0}, \Delta_{2}, \ldots, \Delta_{n-1}$, where the value
$n \in \mathbb{N}$ and the quantities
\[
  \delta_{l} \coloneqq \sum_{k = 0}^{l}
  \frac{\Delta_{k}}{\sum_{j}{\Delta_{j}}}, \quad l \in
  \{0,\ldots,n-1\}
\]
are given.

In order to formulate the Hough transform method for the detection of
those kinds of patterns, we need to define the Hough space and Hough
transform. The domain is
$\mathcal{D} = \{1,\ldots,W\} \times \{1,\ldots,H\}$, where $W$ and
$H$ are width and height of the image. A binary image can be
represented as $\Omega \subset \mathcal{D}$. Now define the Hough
space as $\mathcal{H} = \{1,\ldots,W\}^{\times 2}$. For
$h = (a,b)^{T} \in \mathcal{H}$ the first coordinate defines the
position of the first line and the second coordinate defines the
position of the last line in the pattern. Locations of other lines can
be easily computed, because the relative distances between them are
known. The Hough transform map $\mathcal{P}$ is defined as follows:
\begin{gather*}
  \mathcal{P}(x,y) \coloneqq
  \bigcup\limits_{l=0}^{n-1}
  \left\{
    (a,b) \in \mathcal{H}: x = a + \delta_{l}(b-a)
  \right\}.
\end{gather*}

Let us note, firstly, that the map $\mathcal{P}$ does not depend on the
second coordinate $y$. Therefore, the objective function 
can be rewritten in the form
\begin{gather}
  \sum\limits_{(x,y) \in \Omega} \mathds{1} \left\{ B_{R}(h) \cap
    \mathcal{P}(x,y) \neq \emptyset
  \right\} = \nonumber\\
  \sum\limits_{(x,y) \in \Omega} y \mathds{1} \left\{ B_{R}(h) \cap
    \mathcal{P}(x,y) \neq \emptyset
  \right\}
  \to \max_{h \in \mathcal{H}}.
  \label{eq:2}
\end{gather}

The above problem can also be reformulated as follows.
Instead of considering a binary image $\Omega$, we consider a vector
of length $W$ consisting of tuples with an abscissa coordinate and the
sum of pixel values of the column of the image corresponding to that
abscissa coordinate. This means,
$\bar\Omega = \left\{ (x,y)^{T} \in \{1,\ldots,W\}\times\{1,\ldots,H\}
\right\}$, where $y$ is the $x$-th column sum of the values of the
binary image. The Hough transform map is the same as in the
optimization problem given in \eqref{eq:2}. 
By solving the resulting optimization problem,
 we identify $n$ peaks
in the series $\bar\Omega$ with fixed relative distances
between them. An example of such a series of column means and its pattern of local peaks is shown in Figure~\ref{fig:
  rows sums of parallel lines}.

\begin{figure}[htb]
  \centering
  \begin{tikzpicture}[x=5mm,y=7mm]
    \draw (-1.49,0.3) -- (-1.39,0.86) -- (-1.29,0.06) -- (-1.19,0.65)
    -- (-1.09,0.59) -- (-0.99,0.26) -- (-0.89,0.59) -- (-0.78,0.56) --
    (-0.68,0.61) -- (-0.58,0.87) -- (-0.48,0.15) -- (-0.38,0.81) --
    (-0.28,0.98) -- (-0.18,0.99) -- (-0.08,2.74) -- (0.02,0.69) --
    (0.12,0.75) -- (0.22,0.34) -- (0.33,0.93) -- (0.43,0) --
    (0.53,0.91) -- (0.63,0.94) -- (0.73,0.15) -- (0.83,0.08) --
    (0.93,0.83) -- (1.03,0.3) -- (1.13,0.37) -- (1.23,0.36) --
    (1.33,0.8) -- (1.43,0.46) -- (1.54,0.1) -- (1.64,0.79) --
    (1.74,0.88) -- (1.84,0.43) -- (1.94,2.57) -- (2.04,0.52) --
    (2.14,0.93) -- (2.24,0.08) -- (2.34,0.17) -- (2.44,0.51) --
    (2.54,0.42) -- (2.65,0.06) -- (2.75,0.51) -- (2.85,0.99) --
    (2.95,0.34) -- (3.05,0.5) -- (3.15,0.59) -- (3.25,0.13) --
    (3.35,0.74) -- (3.45,0.34) -- (3.55,0.96) -- (3.65,0.17) --
    (3.75,0.58) -- (3.86,1) -- (3.96,0.89) -- (4.06,0.5) --
    (4.16,0.47) -- (4.26,0.57) -- (4.36,0.38) -- (4.46,2.17) --
    (4.56,0.14) -- (4.66,0.27) -- (4.76,0.87) -- (4.86,0.24) --
    (4.97,0.97) -- (5.07,0.65) -- (5.17,0.81) -- (5.27,0.37) --
    (5.37,0.04) -- (5.47,0.37) -- (5.57,0.33) -- (5.67,0.8) --
    (5.77,0.83) -- (5.87,0.88) -- (5.97,0.41) -- (6.07,0.47) --
    (6.18,0.22) -- (6.28,0.85) -- (6.38,0.34) -- (6.48,0.8) --
    (6.58,0.76) -- (6.68,0.62) -- (6.78,0.46) -- (6.88,0.88) --
    (6.98,0.25) -- (7.08,0.11) -- (7.18,0.95) -- (7.29,0.26) --
    (7.39,0.53) -- (7.49,2.96) -- (7.59,0.1) -- (7.69,0.19) --
    (7.79,0.28) -- (7.89,0.44) -- (7.99,0.8) -- (8.09,0.8) --
    (8.19,0.44) -- (8.29,0.37) -- (8.39,0.6) -- (8.5,0.65) --
    (8.6,0.19) -- (8.7,0.72) -- (8.8,0.64) -- (8.9,0.56) -- (9,0.17);
    \node[above] at (1,2) {$\Delta_{1}$};
    \node[above] at (3.25,2) {$\Delta_{2}$};
    \node[above] at (6,2) {$\Delta_{3}$};
  \end{tikzpicture}
  \caption{The column sums indicate the grid lines.}
  \label{fig: rows sums of parallel lines}
\end{figure}
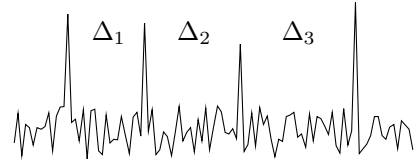


A further improvement can be achieved by adding additional constraints
on the parameter $h \in \mathcal{H}$. Firstly, it is reasonable to
assume that there are two constants $L,U$ such that
$L \leq h_{2} - h_{1} \leq U$, where $h_{1}$ and $h_{2}$ are the two
coordinates of the vector $h$. Those constraints mean that a module
has width more than $L$ and less than $U$ pixels. Secondly, we assume
that the PV module is located completely inside an EL image, such that
$0 \leq h_{1}, h_{2} \leq W$.

Hence, the final formulation of optimization problem is given by
\begin{gather}
  \label{eq:3}
  \begin{split}
    \sum\limits_{(x,y)^{T}\in\Omega}
    y \mathds{1}
    ( B_{R}(h) \cap \mathcal{P}(x,y) \neq \emptyset )
    \to \max_{h \in \mathcal{H}}\\
    L \leq h_{2} - h_{1} \leq U, \quad 0 \leq h_{1},h_{2} \leq W,
  \end{split}
\end{gather}
where $R$ is a method parameter. The parameter $R$ determines the
accuracy of the method by allowing a pattern to fit {\em imprecisely}
inside a binary image in the following sense: The distance between
each line and its true position can be up to $R$ pixels. In our
application it turned out that even small values of $R$ (we chose
$R = 5$) give accurate results. Further, allowing for such an
imprecision automatically provides robustness of the method, when the
pattern lines are slightly distorted by, for example, uncorrected
radial distortion.

We solve the optimization problem using the method described in
Section~\ref{sec:hough-transform}. The most time-consuming step in this
optimization is the evaluation of the transformation map
$\mathcal{P}$. For the present setting, this evaluation boils
down to solving a system of linear inequalities. We list these
inequalities in Appendix A.

In order to apply the Hough Transform approach, introduced above for
an EL image, we need to preprocess it to obtain a binary image. For
the cell detection method we use the well-known Canny edge detection
algorithm implemented in the OpenCV library \cite{2000opencv_library}.

The cell detection method uses other preprocessing method that require
tuning parameters. We choose those tuning parameters in the same
fashion as in Section~\ref{sec:persp-dist-corr} for perspective
distortion correction method, namely, by standardizing the image and
choosing parameters manually.

We summarise the presented method in Algorithm~\ref{algo: cell detection}.

\begin{algorithm}
  \DontPrintSemicolon
  Standardise image; \\
  For a given module type, compute from physical characteristic
  $\delta_{l}$ for vertical and horizontal grid lines; \\
  Apply Canny edge transformation to obtain a binary image;\\
  Perform the Hough Transform optimization twice with the selected
  accuracy parameter $R$ for vertical and horizontal grid lines;\\
  Cut the PV cells according to the detected grid lines.
  \caption{PV module cell detection \label{alg: cell detection}}
  \label{algo: cell detection}
\end{algorithm}

\section{One-Cell-Extraction Based on Change Point Estimation}
\label{sec:one-cell-extraction}

In some cases, an image only contains one cell, for example when
so-called mini modules are used. Such mini-modules are commonly used
in photovoltaic research and development, in order to test new
technologies. In this case, the problem breaks down to the detection
of the cell boundaries to remove the background. Usually, this will
result in a quadrilateral due to the effects discussed in Sections 3
and 4. We propose to correct for this by applying a mapping which maps
a quadrilateral to a rectangle of arbitrary pixel resolution. In this
way, we obtain a standardized cell image.

For the task of cell identification, one can use the following
simplified procedure which detects separately the four corners and
then extracts the associated quadrilateral representing the relevant
cell area.  It uses minimal a priori knowledge about the location of
the cell corners in the sense that one provides four points which are
near the corners but closer to the center of the image, so that the
associated horizontal and vertical lines cross the boundaries of the
cell. We use the pixel values on those lines to detect the boundary,
which is represented as a change-point where the (average) value of
the pixels changes. This observation can be exploited and motivates to
use change-point estimation methods. It is worth mentioning that this
approach also works, if one needs to extract the smallest
quadrilateral whose sides are boundaries of the object of interest,
e.g. if the corners are rounded as it is the case for the
mini-modules, see Figure~\ref{fig: algorithm one cell}, such that they
are more or less virtual.

The details are as follows. Assume that the upper left corner of the
image with resolution $n \times m$ corresponds to the origin $(0,0)$
of the image. To detect the true coordinates
$(x_{0,true} , y_{0,true})$ of the upper left corner we fix a vertical
$y$-coordinate $y_{0}$ provided by the user, which is known to be
larger than the $y$-coordinate, $y_{0,true}$, of the corner. The gray
values of the pixels located on the horizontal row
$(x, y_{0}), x=0,\ldots,n-1$, define a sequence
$X_{0},\ldots,X_{n-1}$, which has a change in the mean, namely at the
index which corresponds to the pixel where the boundary of the cell
area is located. This change-point can be detected by applying the
cumulated sum (CUSUM) change-point estimator, which is described
below. This CUSUM detector outputs $x_{c}$. Next we take some
horizontal $x$-coordinate $x_{0}$, provided by the user, being larger
than the $x$-coordinate, $x_{0,true}$, of the corner. Extract the
pixel values located on the vertical row (line)
$(x_{0},y), y=0,\ldots,m-1$, and denote them
$Y_{0},\ldots,Y_{m-1}$. This sequence has a change-point at the
$y$-coordinate of the boundary, and again we can detect it by applying
the CUSUM detector. It outputs $y_{c}$. The resulting estimate of the
coordinates of the upper left corner is $(x_{c}, y_{c})$. In a similar
way, we can detect the remaining corners.

The CUSUM approach to estimate the change-point works as follows:
Assume it is fed with a sequence, $X_{0},\ldots,X_{n-1}$, assumed to be
a random sample with finite variance, which has a change in its mean
at $q$. This means, we assume that
\[
  E(X_{i}) = \mu_{0}, \quad i < q
\]
and
\[
  E(X_{i}) = \mu_{1}, \quad i \geq q.
\]

Here $\mu_{0}$ denotes the mean of the background pixels and $\mu_{1}$
the mean of the cell area. This model is justified as a kind of a first
order approximation for practical purposes, assuming that the distance
between the average of the background pixels and the average of the
cell area pixels is much larger than the variation of the pixels
within each of those two segments. It can be shown that the
change-point can be consistently estimated by minimizing the objective
function
\[
   RSS(k) = \sum_{i<k}(X_{i}-X_{1:k})^{2}+\sum_{i\geq k}(X_{i}-X_{k:n})^{2}
\]
where $X_{1:k}$ denotes the arithmetic mean of the first $k$
observations and $X_{k:n}$ the arithmetic mean of the remaining data
points. This means, we minimize the residual sum of squares calculated
under the assumption that the change-point is $k$; then the mean of
the pre-change observations is estimated by $X_{k:n}$ and the mean of
observations after the change is estimated by $X_{k:n}$. The term
$\sum_{i<k}(X_{i}-X_{1:k})^{2}$ arising in $RSS(k)$ is a measure of
the variation of the background pixels, as it is $k$ times the sample
variance of those values. Analogously, the second term in $RSS(k)$ is
$(n-k)$ times the sample variance of cell area pixels. By minimizing
$RSS(k)$ we select the best fitting model given the data, and any
value $k^{*}$ minimizing $RSS(k)$ is optimal in this sense. For independent normal
observations this method is the Maximum Likelihood estimator, see \cite{Hawkins1977}.

The output of the above algorithm is in general a quadrilateral. Let
us denote the coordinates of its corners by
$ (x_1,y_1), \dots, (x_4,y_4) $, ordered as upper left, upper right,
lower left and lower right.  Suppose that we want to output an image
of resolution $ W \times H $. Then the
pixel at location $ (x,y) $ of the output image is set to the value of the pixel $ (x', y') $ with $x' $ the largest integer smaller or equal to\\
 $( 1- y^*  ) x_1 +  y^* x_3 + x^* ( ( 1- y^* ) (x_2-x_1) + y^* (x_4-x_3) ) $
and $ y' $ the  largest integer smaller or equal to\\
$( 1- x^*  ) y_1 +  x^* y_2 + y^* ( ( 1- x^* ) (y_3-y_1) + x^* (y_4-y_3) ) $.
where $ x^* = x / W $ and $ y^* = y/H $. This transformation maps the
quadrilateral to the target rectangle and interpolates in between.

\section{Application and Simulations}
\label{sec:simulations-1}

In this section, we apply the proposed methods to real EL images and
report about extensive computer simulations.  The simulations study
the accuracy of the preprocessing method discussed in
Sections~\ref{sec:rotat-dist-corr} - \ref{sec:grid-line-detection},
where errors propagated through the stages may aggregate. In a first
Monte-Carlo study we sample from our database of real EL images and,
for each sampled image, simulate distortions due to rotation,
perspective and shift. In a second computer simulation we illustrate
the sensitivity with respect to image quality in the sense of
contrast. Here, for two selected real EL images of low and high
contrast the accuracy of the preprocessing work-flow is simulated.

\subsection{Application}


We apply our methods to a collection of approximately 2000 EL
images. Those images were collected under outdoor (field) conditions in several
solar parks. They are taken with a special EL camera that is able to
capture the emitted light spectrum from a PV module. The images are
stored in a JPEG format and have resolution of approximately
$4200\times2800$ pixels.

\begin{figure*}[htb]
  \centering
  \begin{minipage}{.49\textwidth}
    \includegraphics[width=\textwidth]{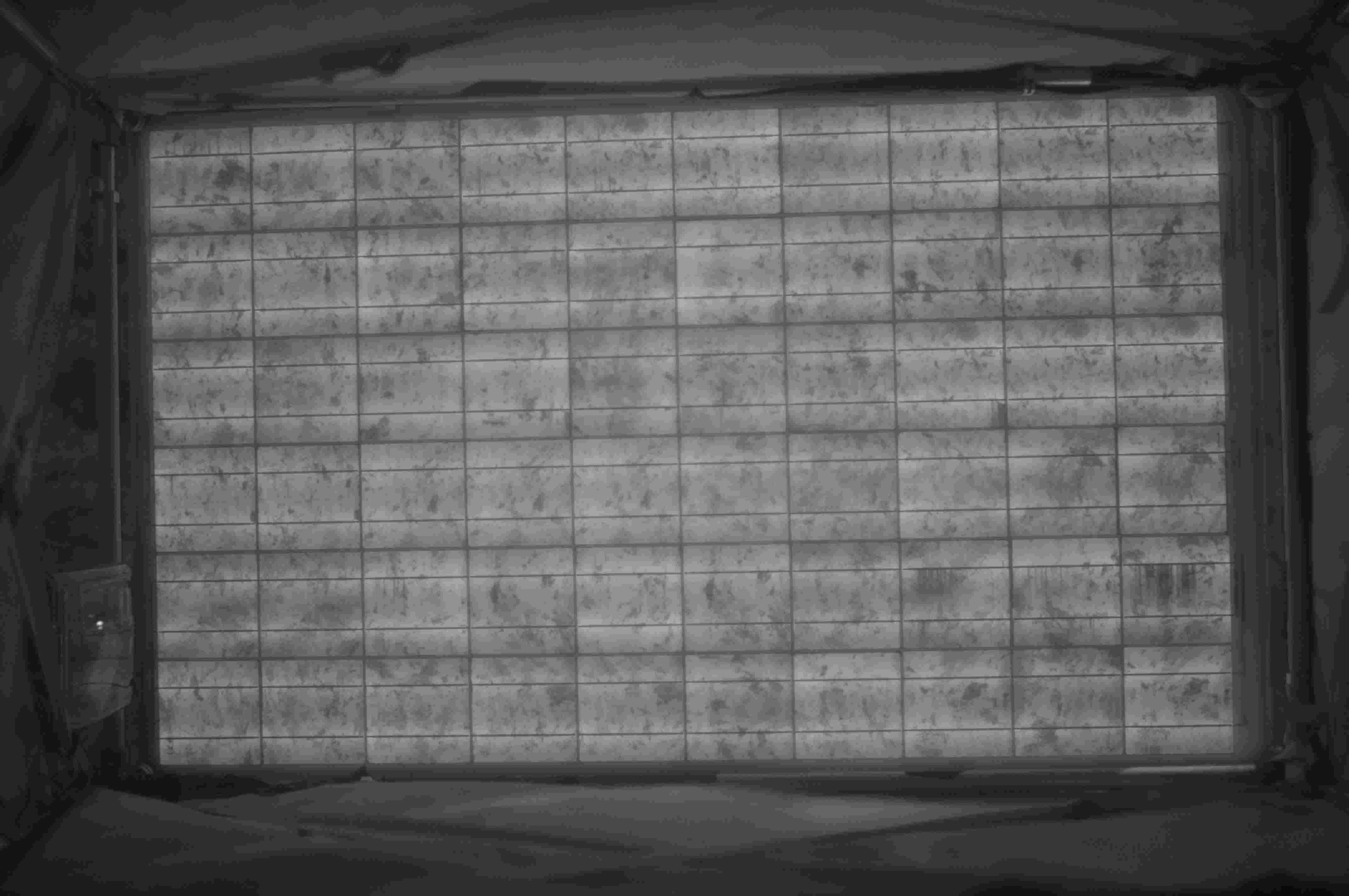}
  \end{minipage}
  \hfill
  \begin{minipage}{.49\textwidth}
    \includegraphics[width=\textwidth]{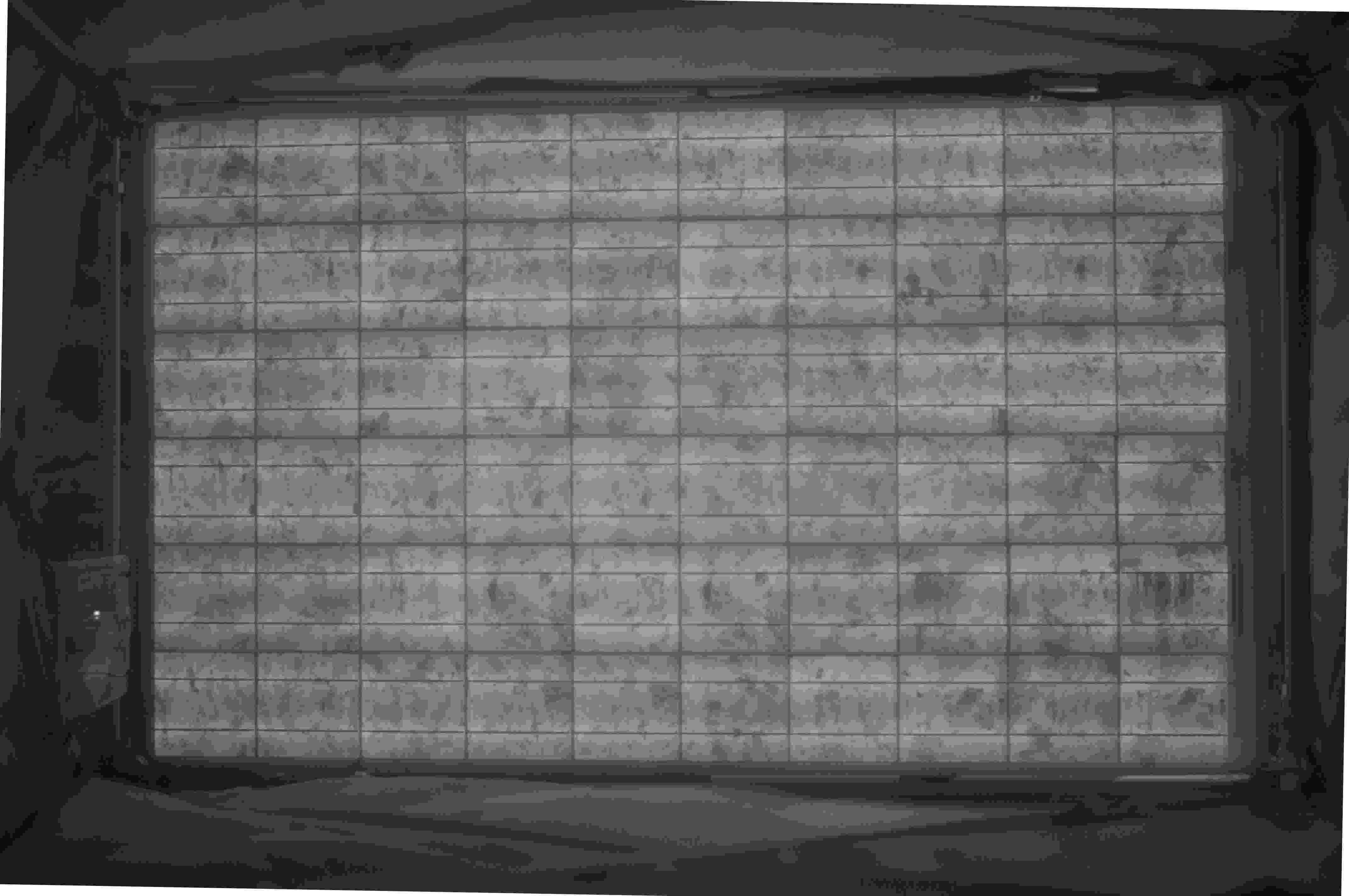}
  \end{minipage}
  \caption{An example of applying rotation correction algorithm. The
    left image is the original EL image. The right image is the method
    output.}
  \label{fig:rotation before and after}
\end{figure*}

Figure~\ref{fig:rotation before and after} shows the output (right
image) of the rotation distortion correction method applied to an EL
image (right image). Note that the method outputs a rotated image
where all vertical grid lines are parallel to the image edges. Among
2000 EL images in our data set there were no images founds, for which
the method did not perform visually correctly. However, as one can
observe in Figure~\ref{fig:rotation before and after}, the horizontal
grid lines are not yet perfectly parallel to the edges of the image
due to the perspective distortion.

\begin{figure*}[htb]
  \centering
  \begin{minipage}{.48\textwidth}
    \includegraphics[width=\textwidth]{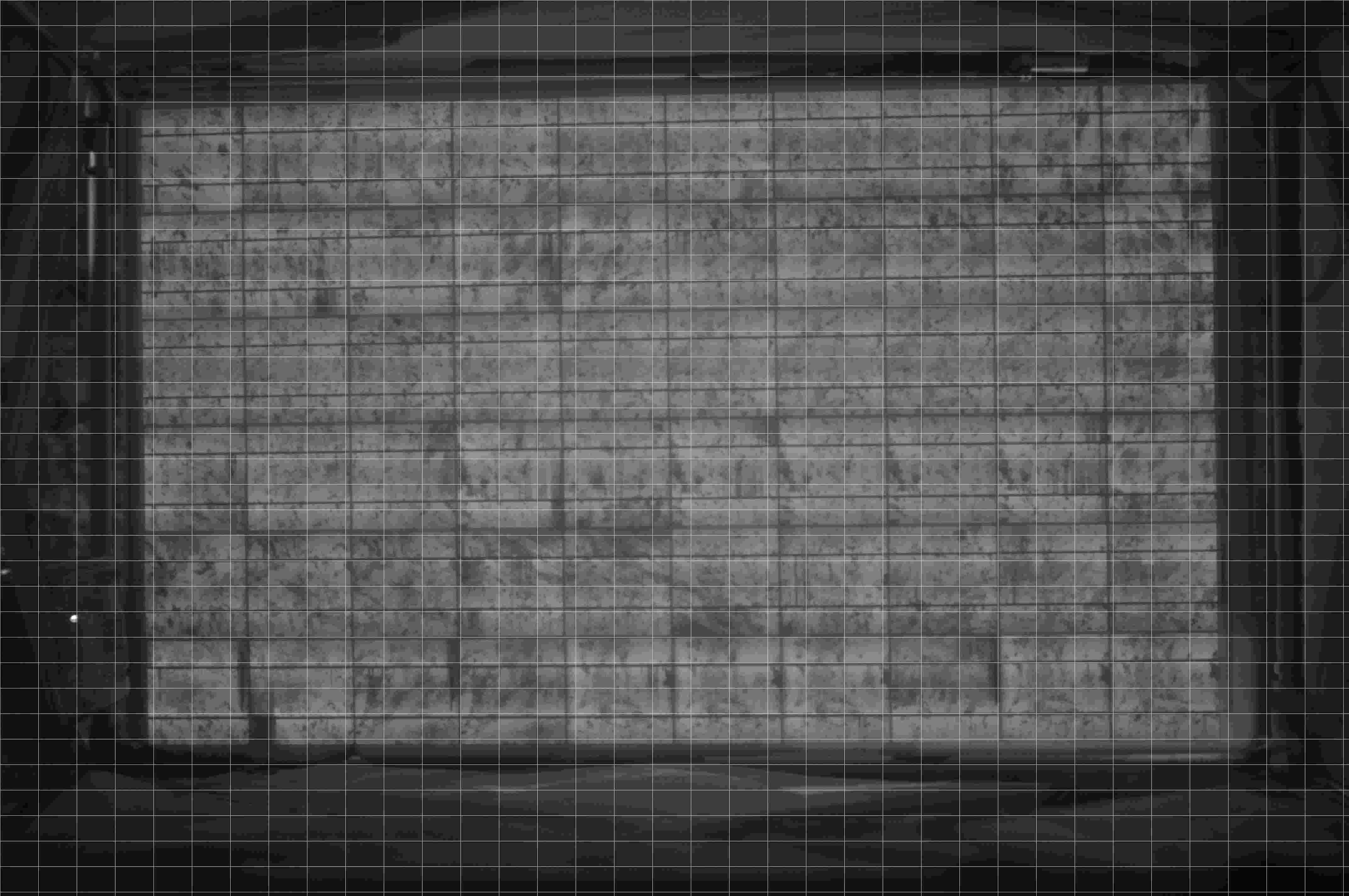}
  \end{minipage} \hfill
  \begin{minipage}{.48\textwidth}
    \includegraphics[width=\textwidth]{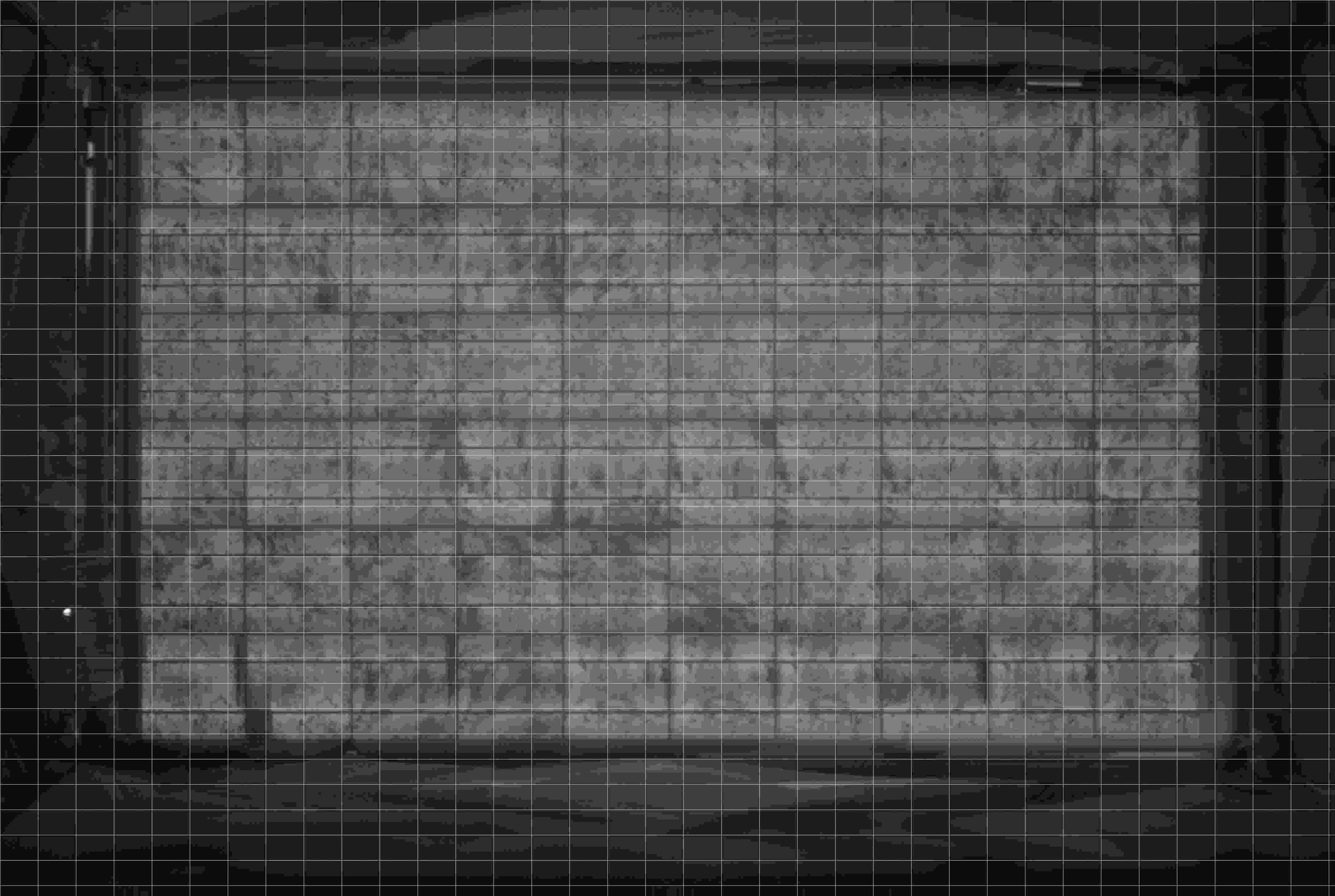}
  \end{minipage}
  \caption{Outdoor EL images are affected by perspective distortion
    (left image), which needs to be corrected in the presence of
    possibly disturbing background. The right image shows the output
    of the proposed method. The white grid lines are added to ease
    visual evaluation of the result.}
  \label{fig: algorithm work examples 2}
\end{figure*}

The application of the perspective distortion correction method is
illustrated in Figure~\ref{fig: algorithm work examples 2}. The white
grid lines are added to assist the visual verification of the achieved
distortion correction. Among the 2000  EL images we analyzed, there are only
a few images for which the method did not work correctly, since only
one vertical or horizontal line was detected, which is not sufficient
to estimate the perspective distortion parameters.

The perspective distortion correction method relies on image
processing sub-procedures that have several tuning parameters. The
USAN edge detector has three parameters and we select them to be equal
$H = 5$, $h = 3$ and $p = 0.4$ (see notations in
\cite{2007Rafajlowicz}). The Hough Line Transform we use has two
parameters: a line threshold equals $50$ and the maximum gap between
points forming a line equals $75$.

\begin{figure*}
  \centering
  \begin{minipage}{.48\textwidth}
    \includegraphics[width=\textwidth]{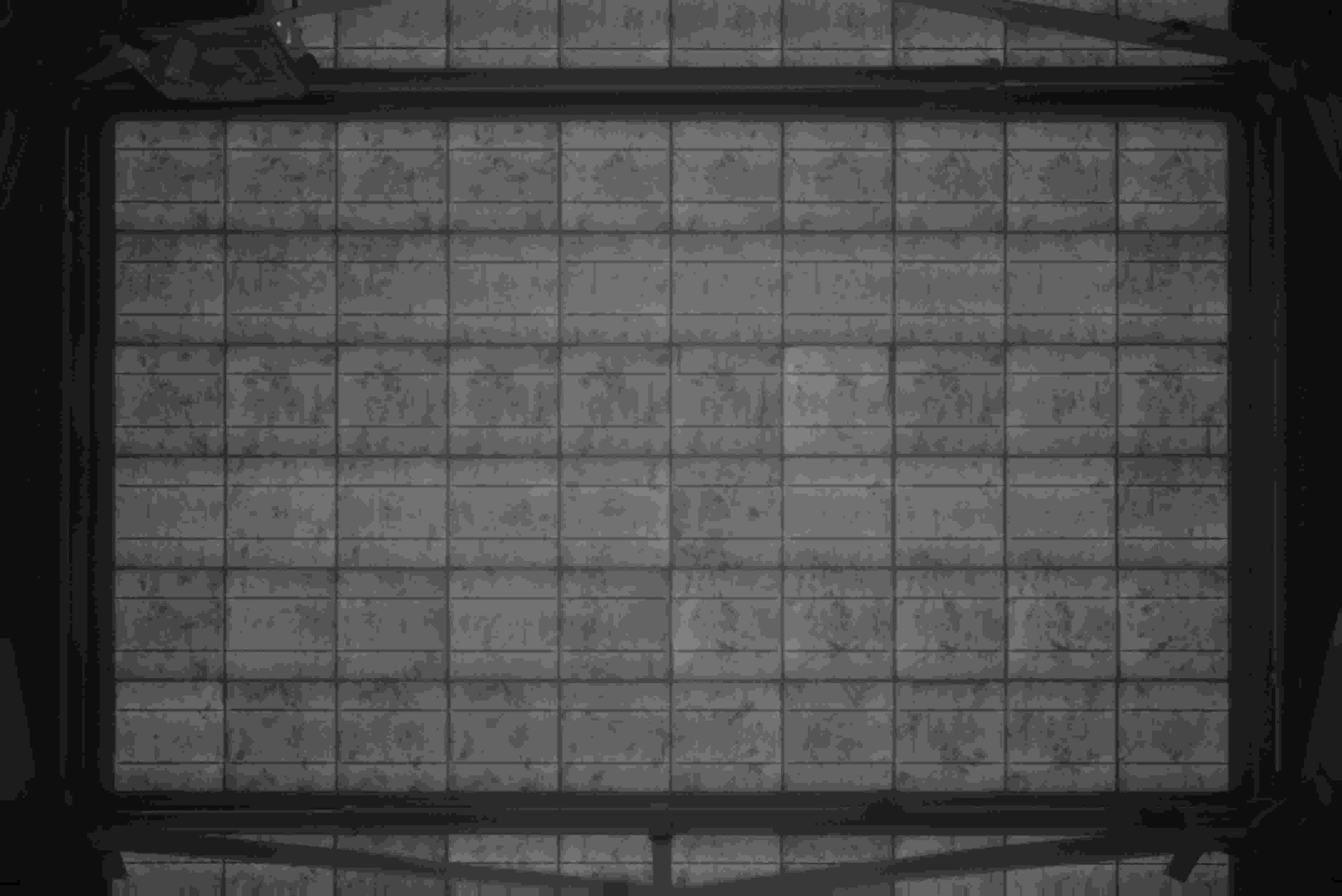}
  \end{minipage}
  \begin{minipage}{.48\textwidth}
    \includegraphics[width=\textwidth]{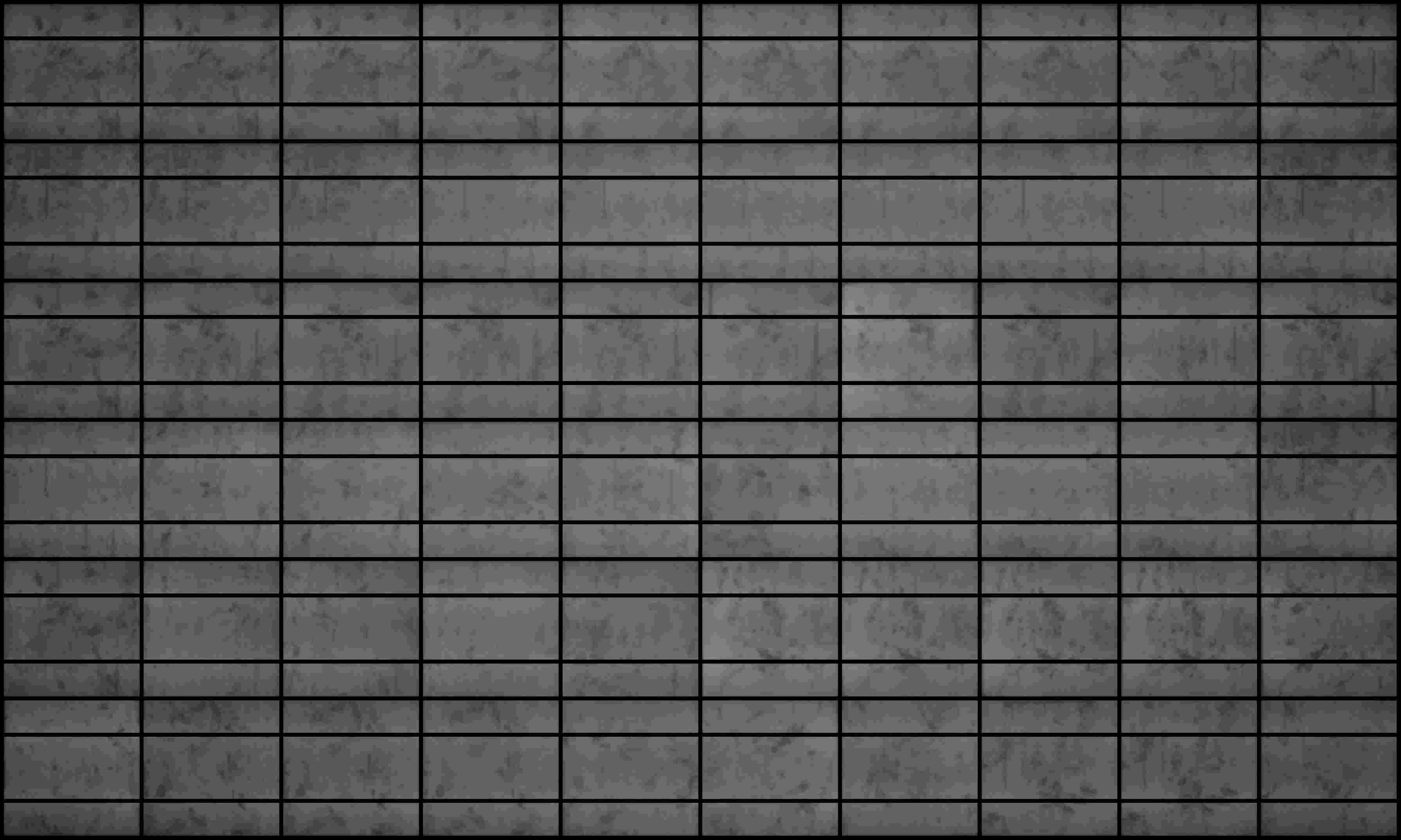}
  \end{minipage}
  \begin{minipage}{.48\textwidth}
    \includegraphics[width=\textwidth]{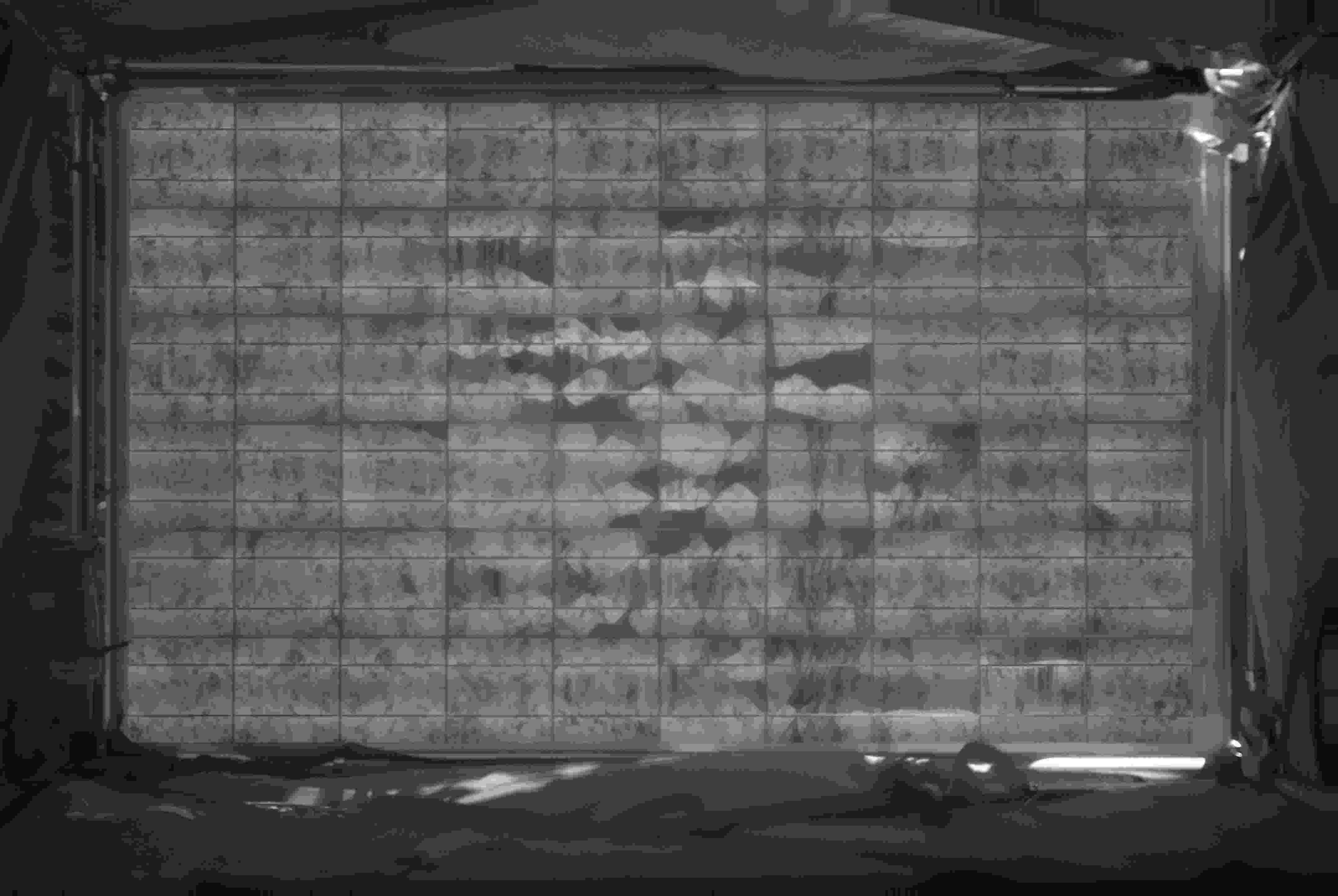}
  \end{minipage}
  \begin{minipage}{.48\textwidth}
    \includegraphics[width=\textwidth]{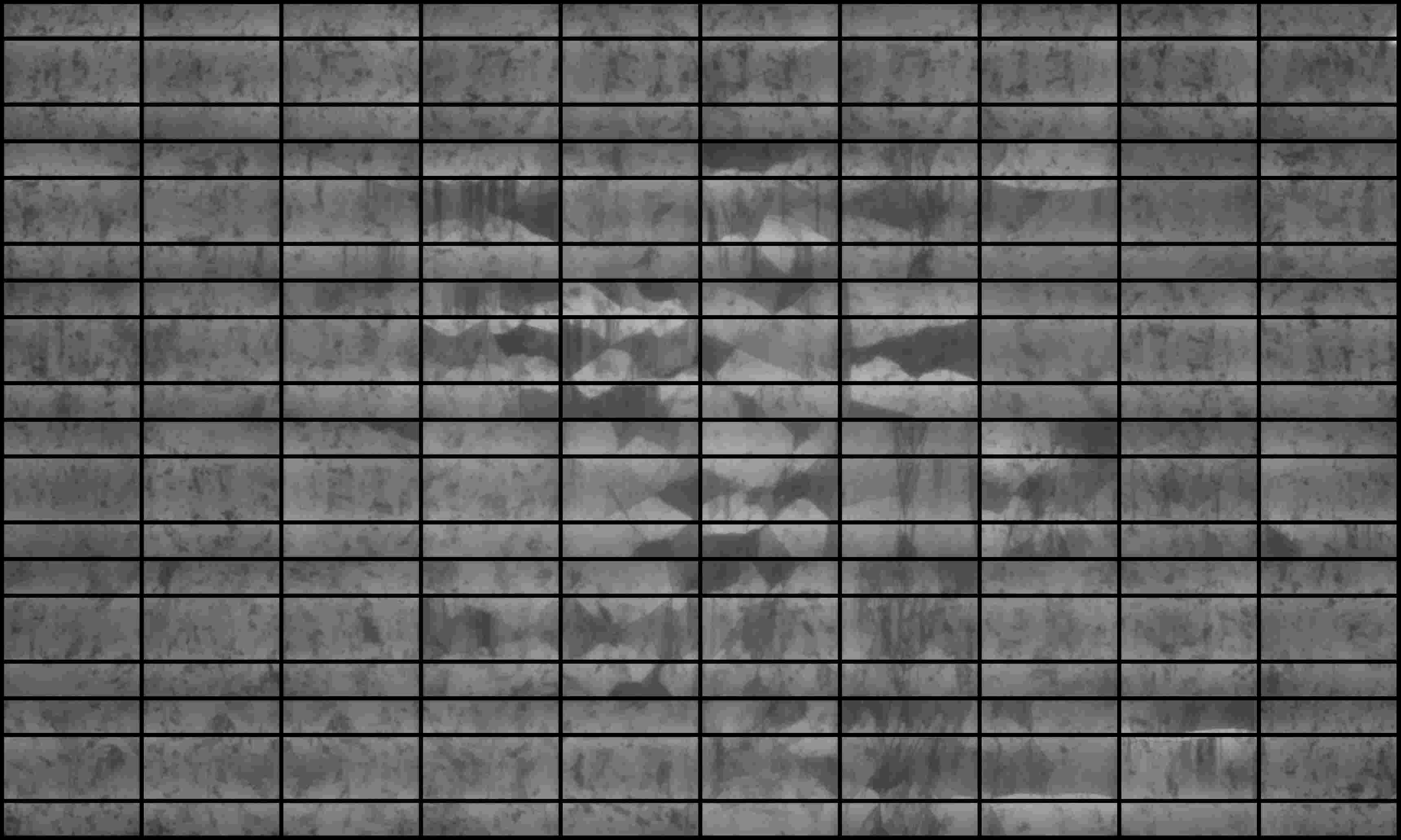}
  \end{minipage}
  \begin{minipage}{.48\textwidth}
    \includegraphics[width=\textwidth]{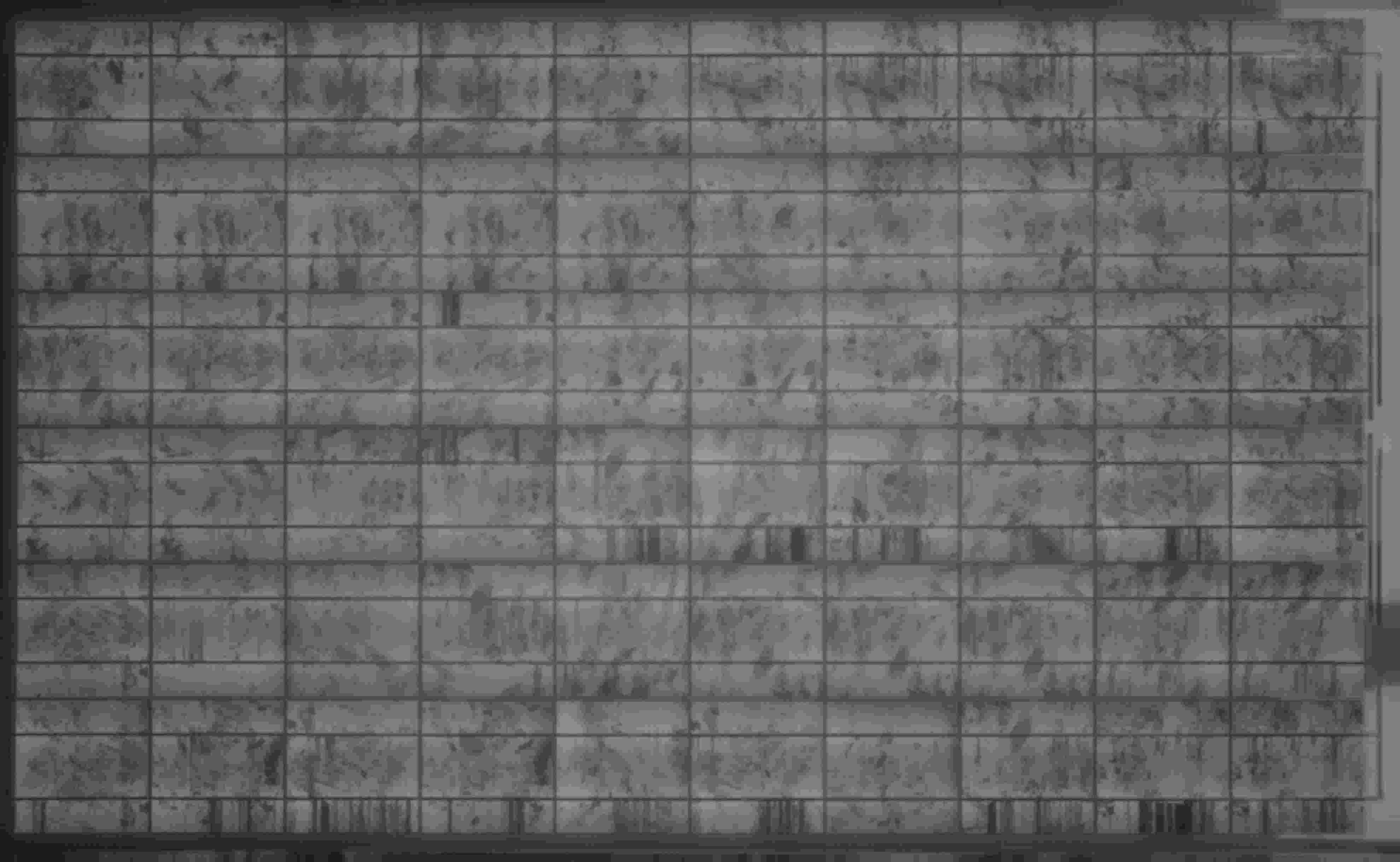}
  \end{minipage}
  \begin{minipage}{.48\textwidth}
    \includegraphics[width=\textwidth]{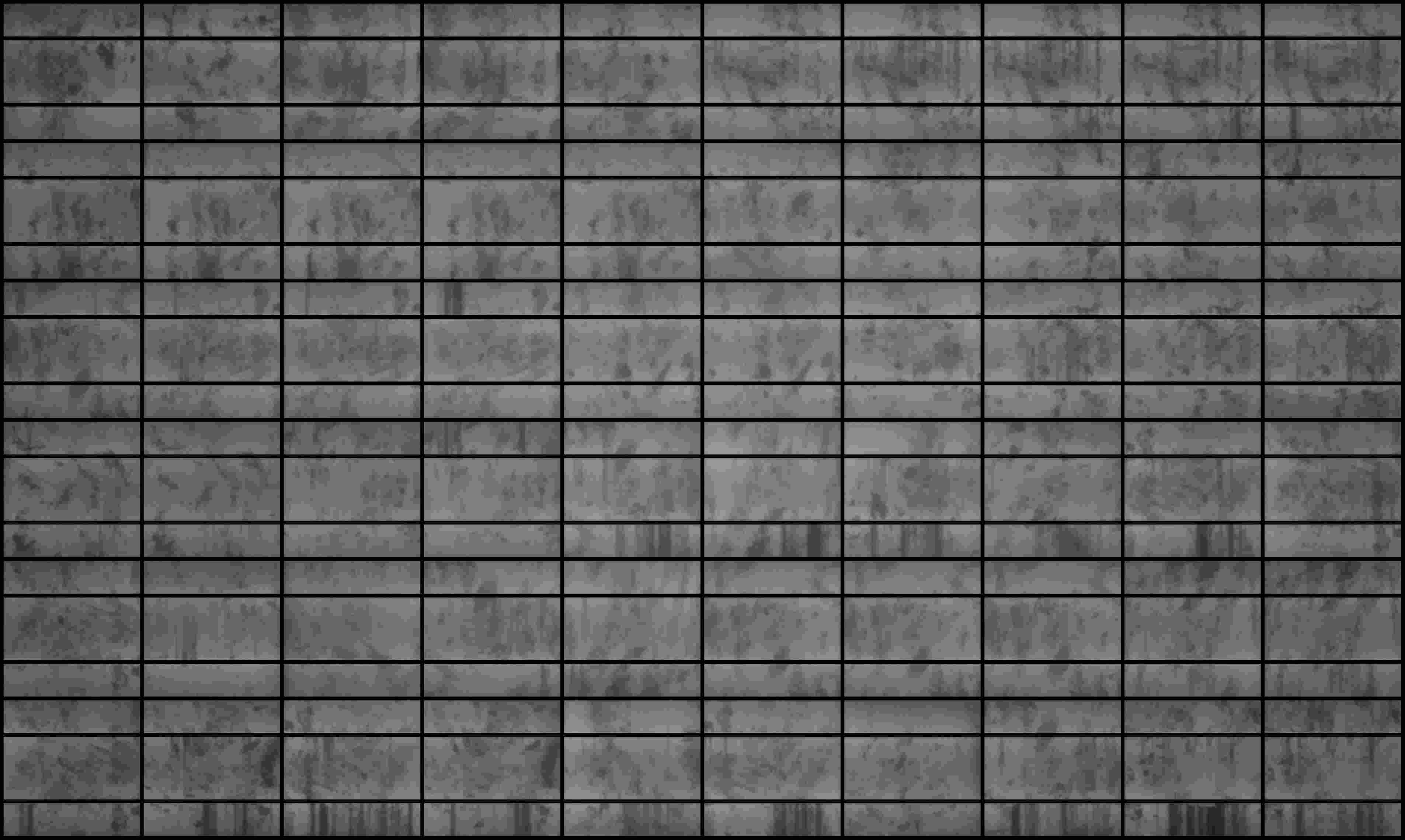}
  \end{minipage}
  \caption{Examples of cell detection algorithm. The left images are
    EL images with corrected perspective distortion. The right images
    are images with detected PV cells combined together to form the
    original PV module}
  \label{fig: examples module detection}
\end{figure*}

Figure~\ref{fig: examples module detection} shows the application of
the cell detection method to several preprocessed EL images. We also
applied it to our data base. Among 2000 sample EL images there were
only a couple of images where the method did not work correctly, as a
consequence of low contrast of the image. It is worth noting that the
method performs well even when not all cells in the module are
connected to a circuit or if they are heavily damaged.

The cell detection method uses Canny edge detector, a procedure that
has three tuning parameters. The lower threshold parameter is chosen
to be equal 25, the upper threshold parameter equals 75 and the kernel
size equals 3. Furthermore, the Hough transform has itself a single
tuning parameter $R = 5$. The (relative) distances between the grid
lines have been measured by hand, as in our EL images database there
are only 4 different types of modules.

\begin{figure*}[htb]
	\centering
	\begin{minipage}{.48\textwidth}
		\includegraphics[width=\textwidth]{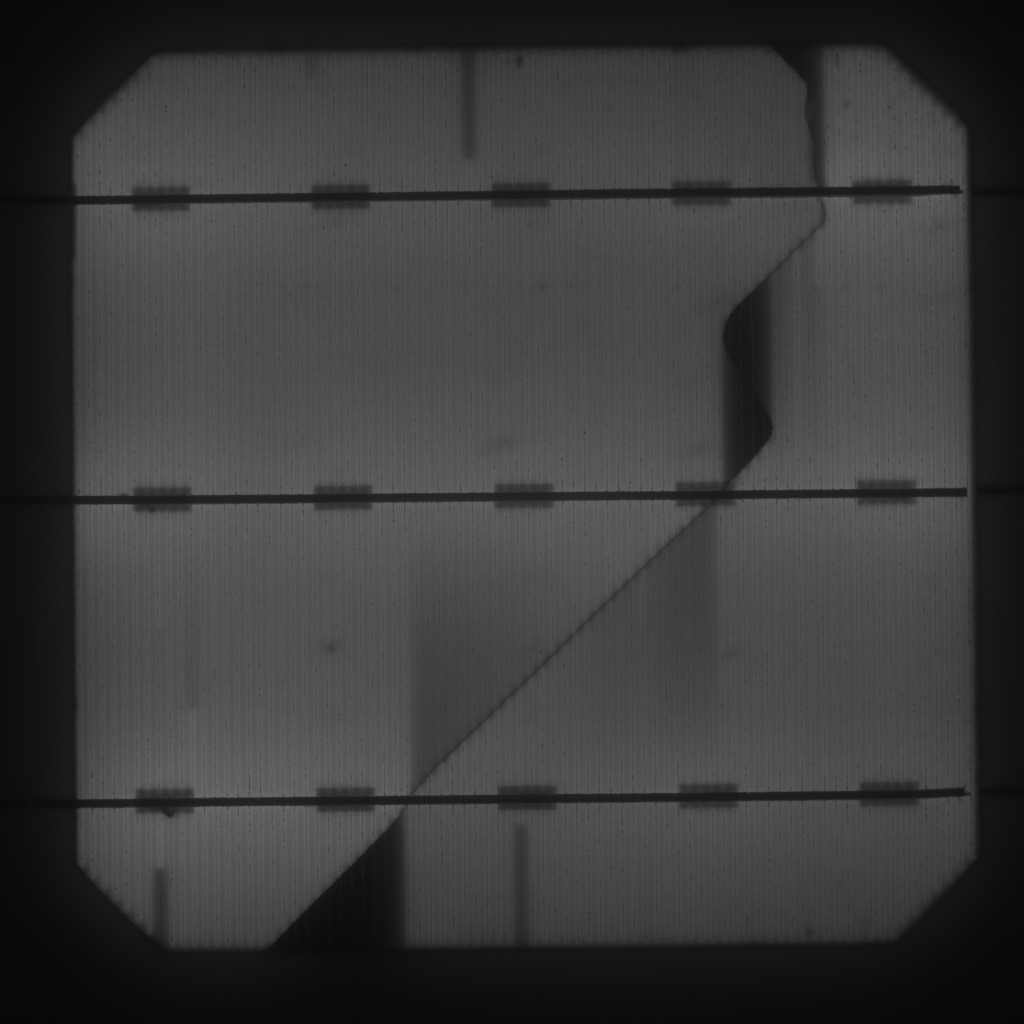}
	\end{minipage} \hfill
	\begin{minipage}{.48\textwidth}
		\includegraphics[width=\textwidth]{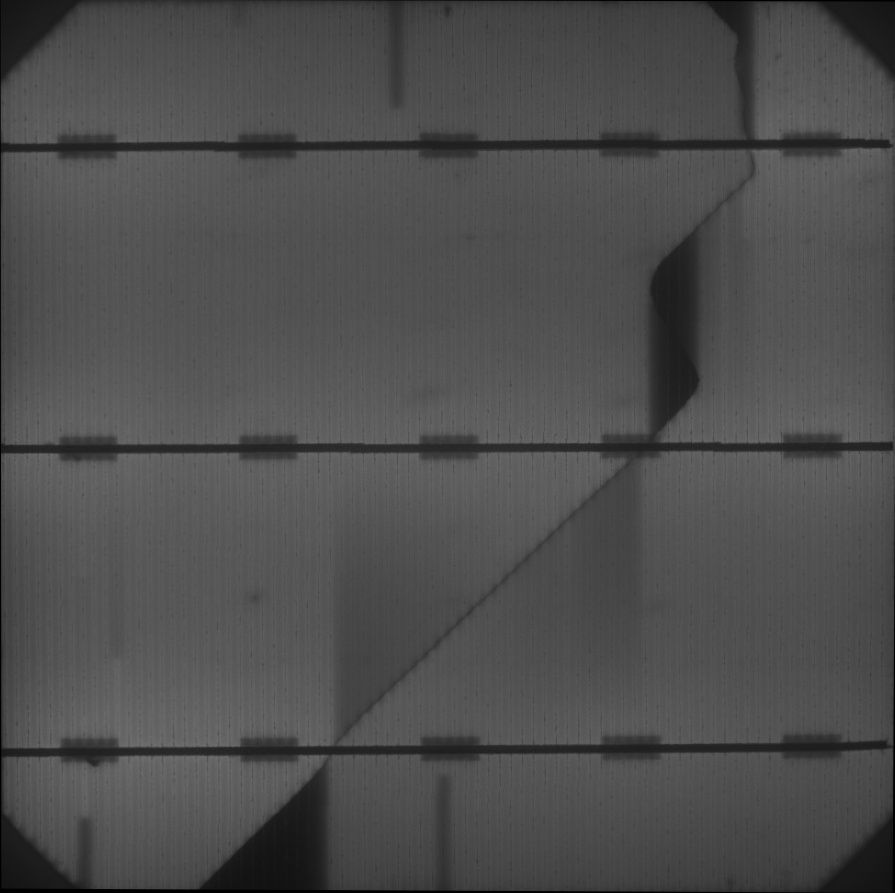}
	\end{minipage}
	\caption{EL image of a mini-module (left) and the extracted and
		rescaled cell area using the one-cell extraction method.}
	\label{fig: algorithm one cell}
\end{figure*}

Lastly, Figure~\ref{fig: algorithm one cell} shows the application of
the simplified, fast CUSUM procedure to extract the solar cell from a
mini-module. It can be seen that the approach works reliable and
extracts the solar cell with high accuracy.

\subsection{Database simulation}

To assess the performance of the proposed methods we employ the
following scheme. First, all images of the database were corrected
using the proposed preprocessing work-flow. The resulting corrected
database was taken as ground truth. Next, a random sample of $ 100 $
images was drawn. To each sampled image simulated random rotation,
perspective and shift distortions were applied by drawing the
characterizing parameters from probability distributions. The details
of this parameterization and the choice of the distributions are given
below.  For every type of distortion (rotation, perspective, shift)
$100$ different parameters resp. parameter vectors were drawn
resulting in a total of $ 10^4 $ images to which the preprocessing
procedure was applied.

The accuracy of the method is measured by comparing the known
(simulated) parameter vector, $ \theta $, of distortion parameters
with the estimated one, $ \widehat{\theta} $, calculated from the
preprocessing output image. As a measure we use the sum of absolute
deviations (Manhattan distance),
$$ SAD = \sum_i |\theta_i - \widehat{\theta}_i|, $$ separately
calculated for rotation, perspective, module position and module
size. In total, this simulation scheme provides $ 10^4 $ values of
each SAD, and these distributions were analyzed.

Before discussing the results, let us describe the details of the
simulation step:

The {\em rotation distortion} is governed by a single parameter, the
rotation angle, which is drawn from a uniform distribution on the
interval $[-20,20]$ (measured in degrees).

The {\em perspective distortion} is governed by $8$ parameters, namely
the coordinates of the distorted rectangle in Figure~\ref{fig:
  correcting rectangular}). We sample those $8$ parameters uniformly
in a hyper-interval in such a way that the variation of the boundary
lines of the rectangular PV module is approximately $ \pm 5 $ degrees.

Lastly, the {\em shift} was simulated with periodic boundary
conditions, i.e. on a torus, in such a way that the PV module lies
within the original image boundaries. This shift is parameterized by a
two-dimensional vector: shifts along $x$ and $y$ axes. In addition, we
considered the associated module size as a parameter.

It is worth mentioning that the ranges of the uniform distributions
used here to simulate distortion effects lead to stronger distortions
than present in our database of real EL images. These images were
collected under outdoor conditions without demounting the PV modules,
and therefore one can expect even higher accuracy when using the
approach for image data collected under well-controlled conditions or
in a lab.

After applying the simulated distortion effects, the preprocessing
work-flow including the cell detection stage was run. Then the estimated
parameter vector $ \widehat{\theta} $ and its Manhattan distance to
$ \theta $ were determined. For the shift parameters the calculations
are somewhat involved: For a PV module the distance between its
top-left corner $x$ coordinate on the original module (before
applying the random distortions) and its top-left corner $x$
coordinate on the output image (after applying preprocessing and cell
detection) were calculated. Adding the corresponding distance of the
$y$ coordinates gives the value of the SAD. At this point, it is worth
recalling that the cell detection method was applied with accuracy
parameter $R=5$.

\begin{figure*}
  \centering
  \begin{minipage}{.26\linewidth}
    \includegraphics[width=\textwidth]{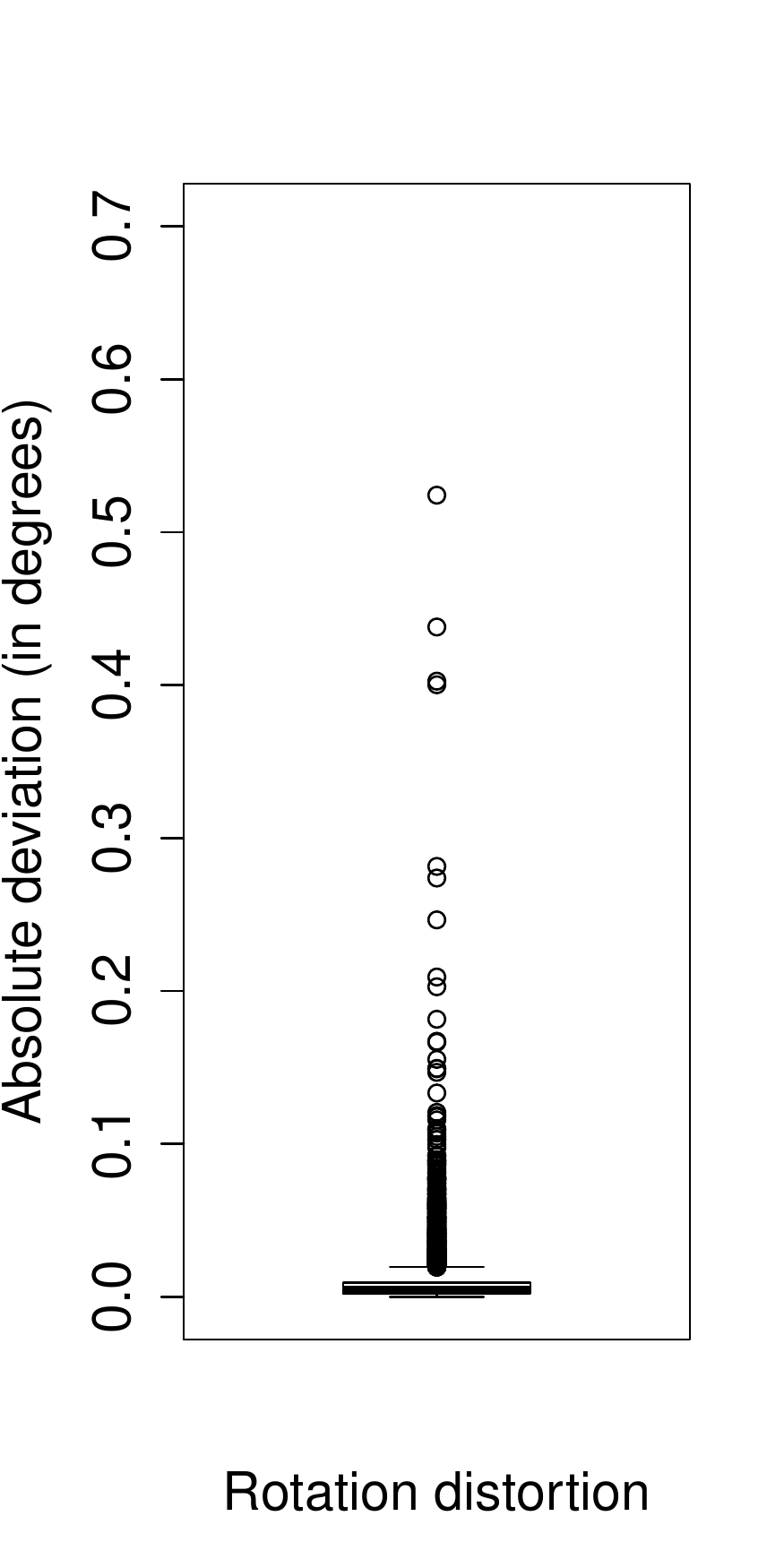}
  \end{minipage}
  \begin{minipage}{.26\linewidth}
    \includegraphics[width=\textwidth]{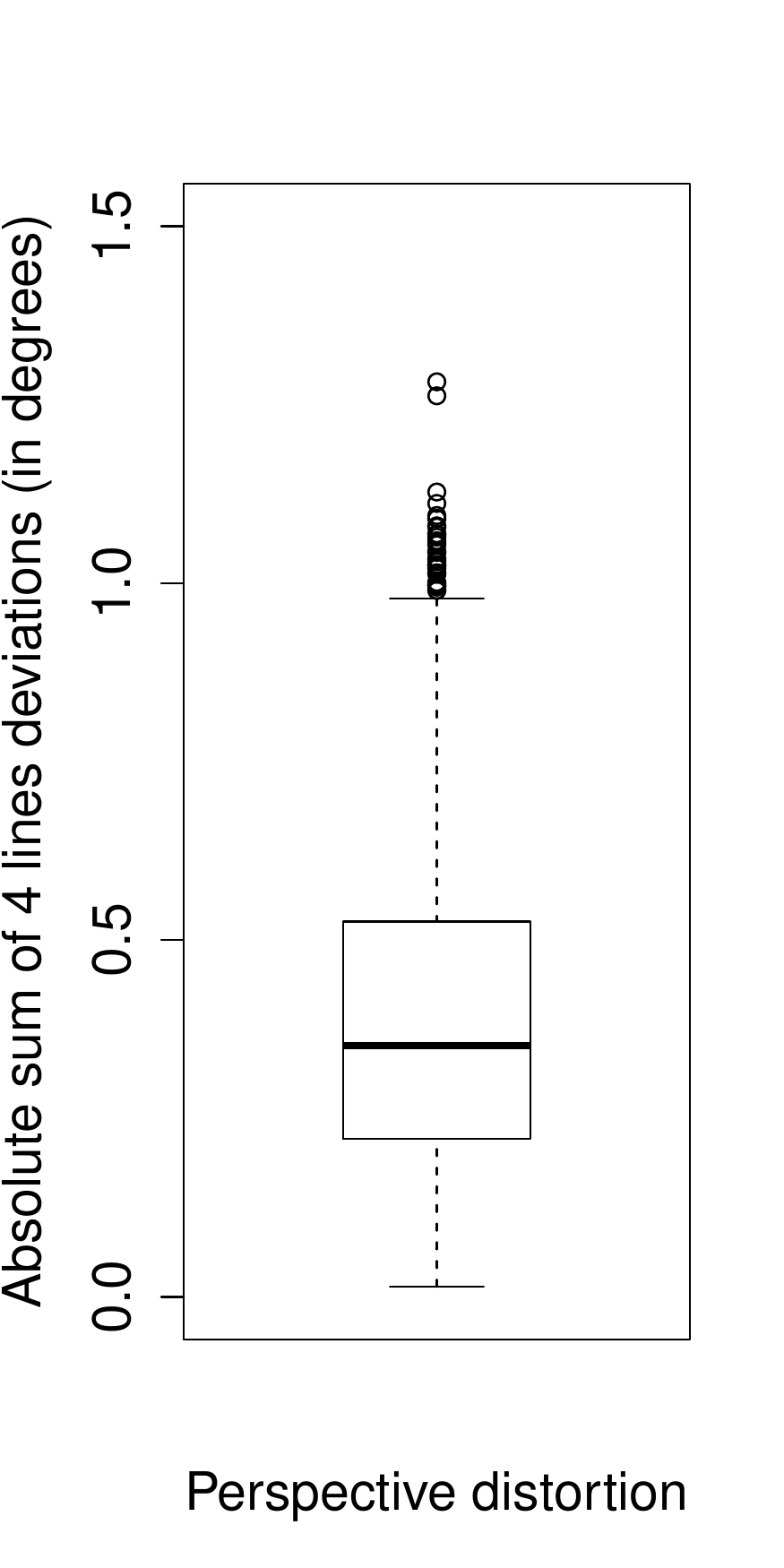}
  \end{minipage}
  \begin{minipage}{.365\linewidth}
    \includegraphics[width=\textwidth]{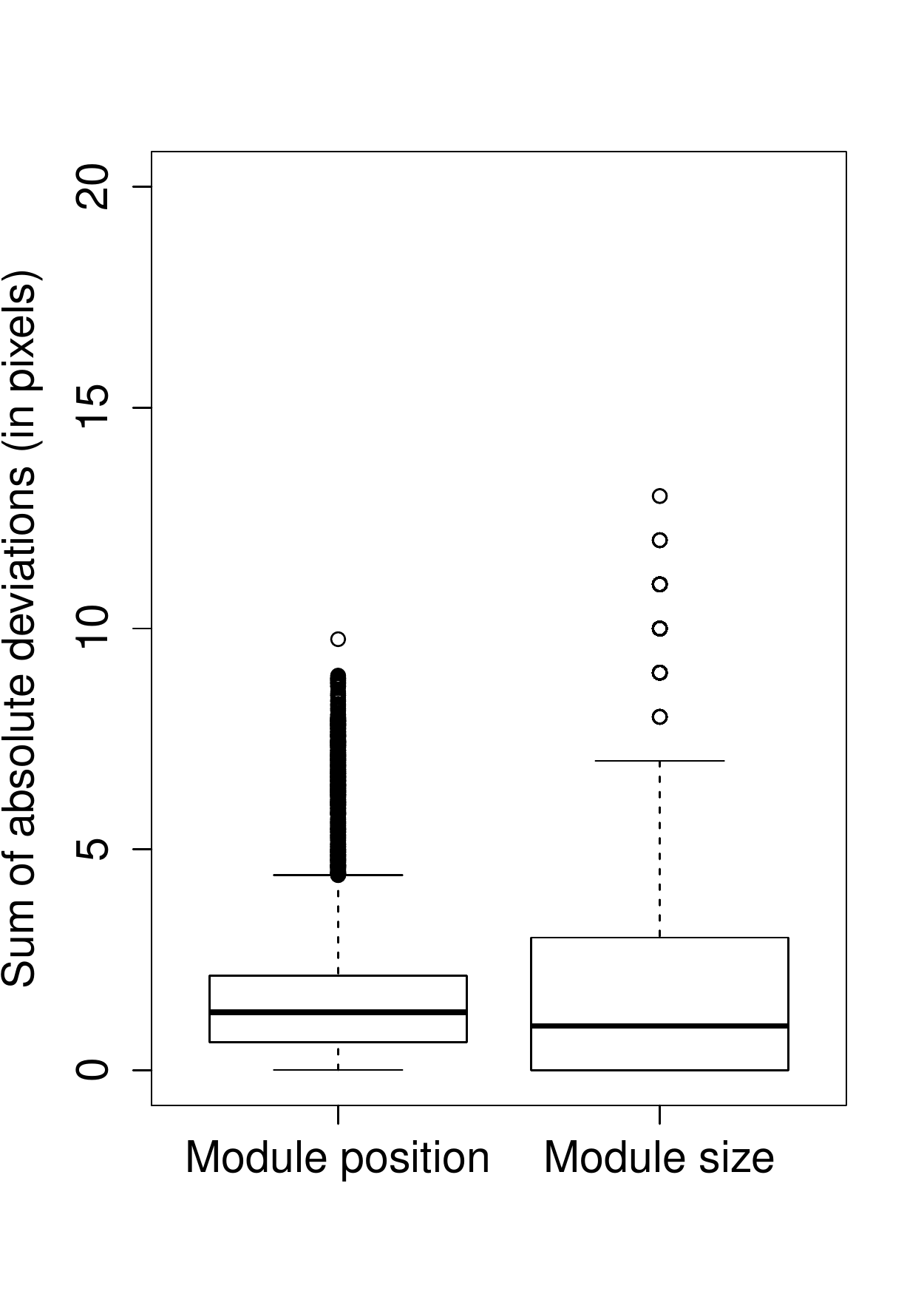}
  \end{minipage}
  \caption{Performance of the preprocessing methods for the database. Left to right:
    rotation distortion correction, perspective distortion correction
    and cell detection accuracy}
  \label{fig: preprocessing performance}
\end{figure*}

The resulting distribution of the accuracy measured by SAD is depicted
in Figure~\ref{fig: preprocessing performance}, where boxplots for the
SAD of rotation, perspective, module position and module size are
shown.  The results demonstrate the high accuracy of the approach. For
example, the measure for perspective distortion, the sum of the
distances (in degree) of all four sides, is less than 1 degree except
a few outliers and not larger than $0.6$ degree for $3/4$ of the
cases.

We also remark that all preprocessing methods are applied with a fixed
set of tuning parameters. The simulations not presented here indicate
that there is almost no variation in performance for different EL
images, and therefore, the selected set of tuning parameters adapt
well for images in our database. However, there are few cases of
images with relatively low visual contrast where a small difference in
performance can be noticed. We address this issue in the next section.

\subsection{Simulations for low and high contrast EL images}

In order to investigate how the proposed approach performs depending
on the image quality of an EL image, we selected two images from the
database which are shown in Figure~\ref{fig: high and low contrast el
  images}. On a high contrast image (right image in Figure~\ref{fig:
  high and low contrast el images}) the grid lines are much better
reproduced than on a low contrast image (left image).  The selected
low contrast image also suffers from severe vignetting, i.e. darkening
of the corners, and it is somewhat underexposed.  The inhomogeneity of
the cell area, due to the multicrystalline silicon of that module, is
also more pronounced than for the high contrast image. The question
arises to which extent this has an effect on the accuracy of our
method.

\begin{figure*}[htb]
  \centering
  \begin{minipage}{.48\textwidth}
    \includegraphics[width=\textwidth]{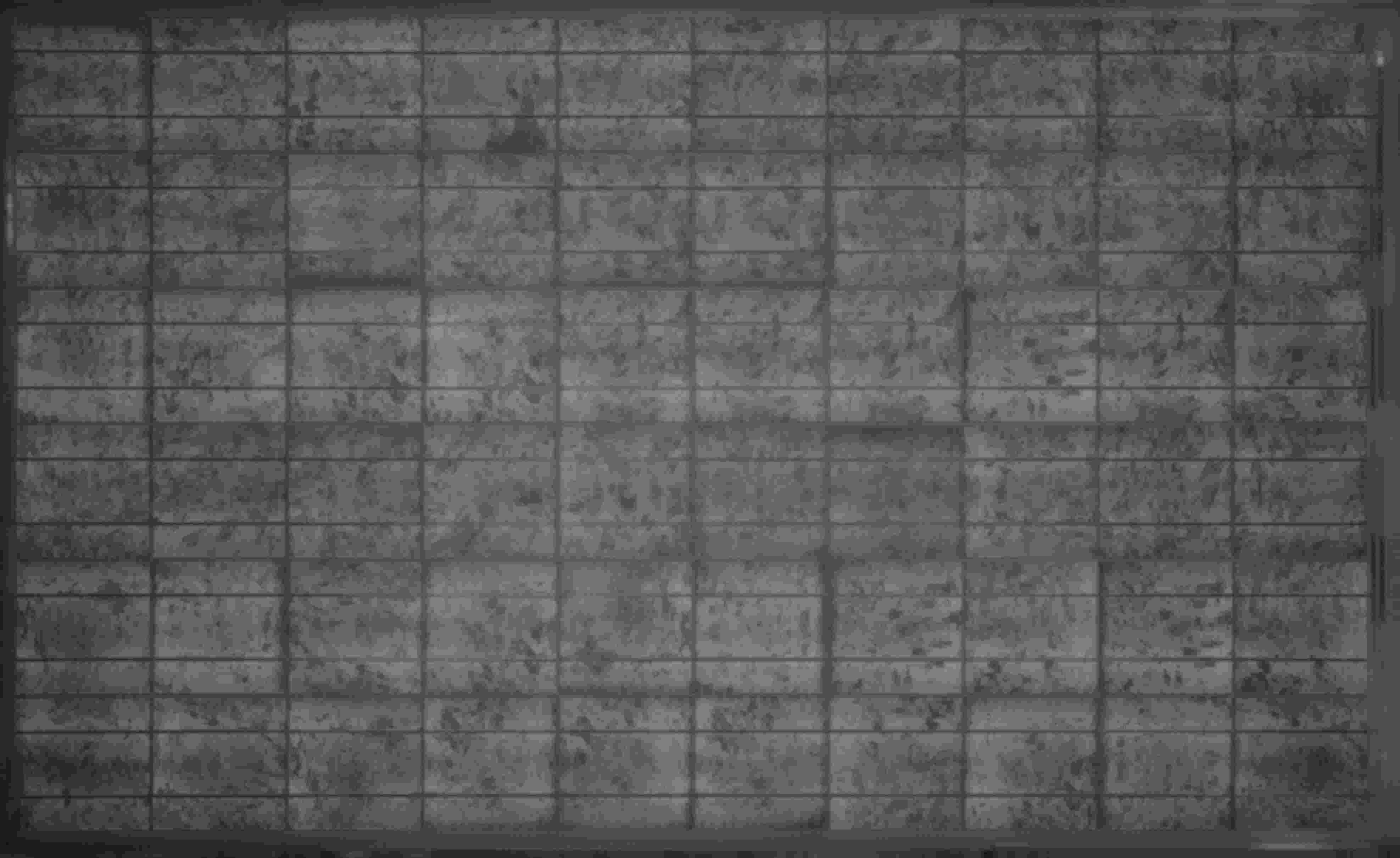}
  \end{minipage} \hfill
  \begin{minipage}{.48\textwidth}
    \includegraphics[width=\textwidth]{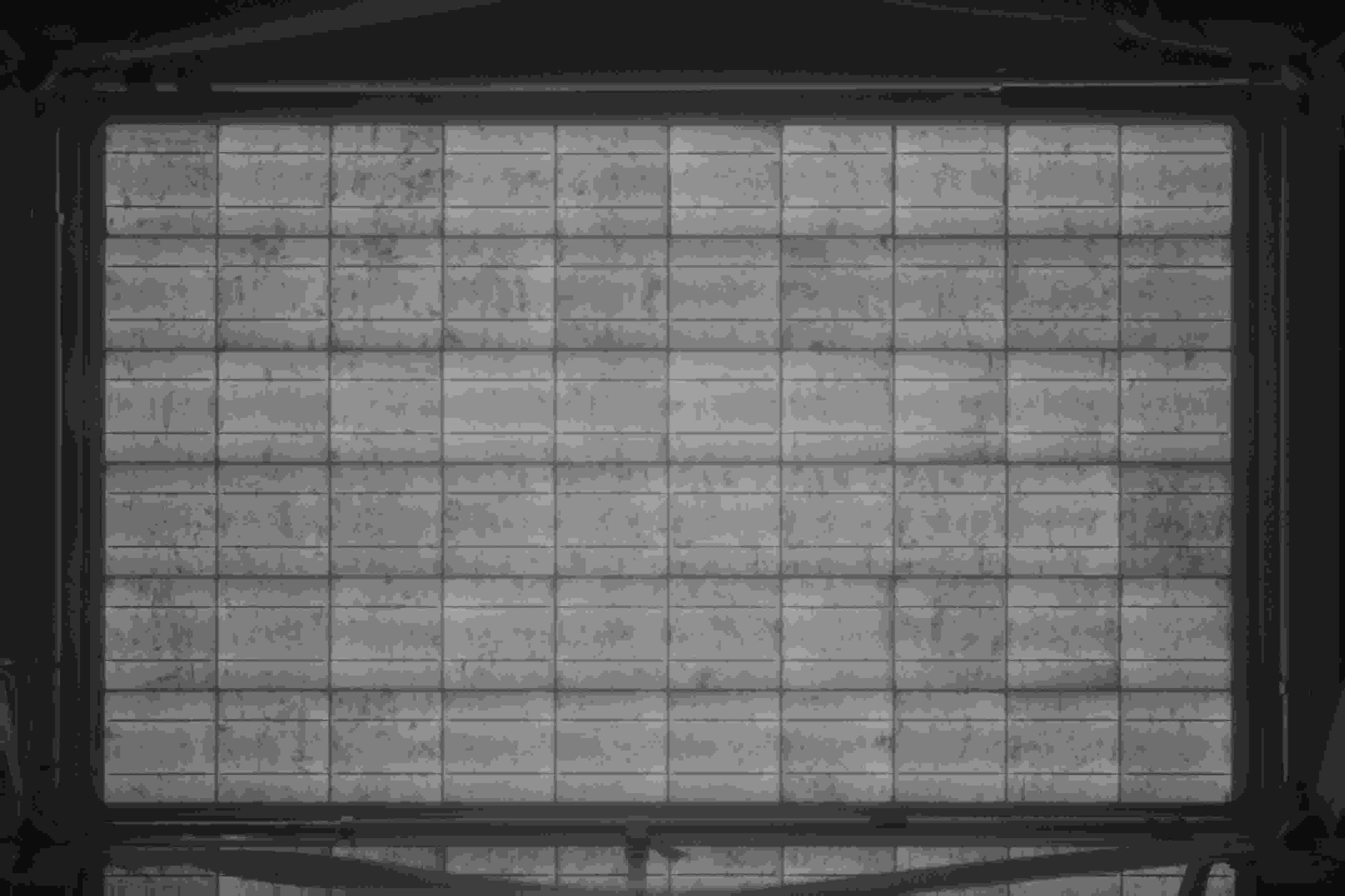}
  \end{minipage}
  \caption{An example of two images with low (left image) and high
    (right image) contrast}
  \label{fig: high and low contrast el images}
\end{figure*}

The simulation scheme discussed above in detail was run for both
images providing two distributions of sum of absolute differences
(SAD) measure for each of the effects (rotation, perspective, module
position, module size). The result is shown in Figure~\ref{fig: high
  low contrast performance}. It can be seen that image quality in
terms of low contrast has an impact, but the accuracy is still very
good. We may conclude that the proposed preprocessing framework is
capable of dealing with low image contrast EL images.  As a
consequence, when mainstreaming the approach in an industrial
production, one can expect reliable, robust and accurate performance
even under high-throughput conditions.

\begin{figure*}
  \centering
  \begin{minipage}{.245\linewidth}
    \includegraphics[width=\textwidth]{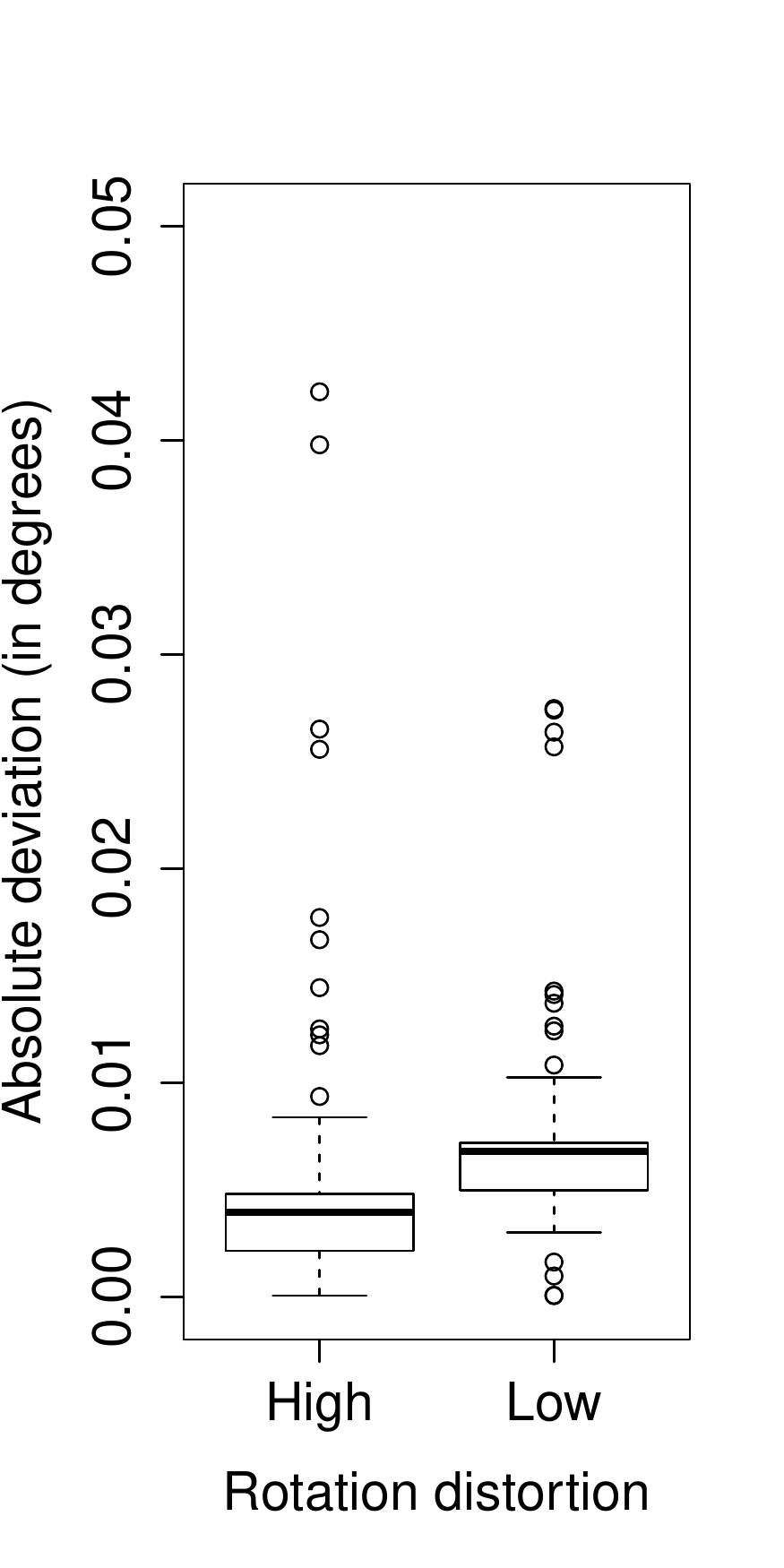}
  \end{minipage}
  \begin{minipage}{.245\linewidth}
    \includegraphics[width=\textwidth]{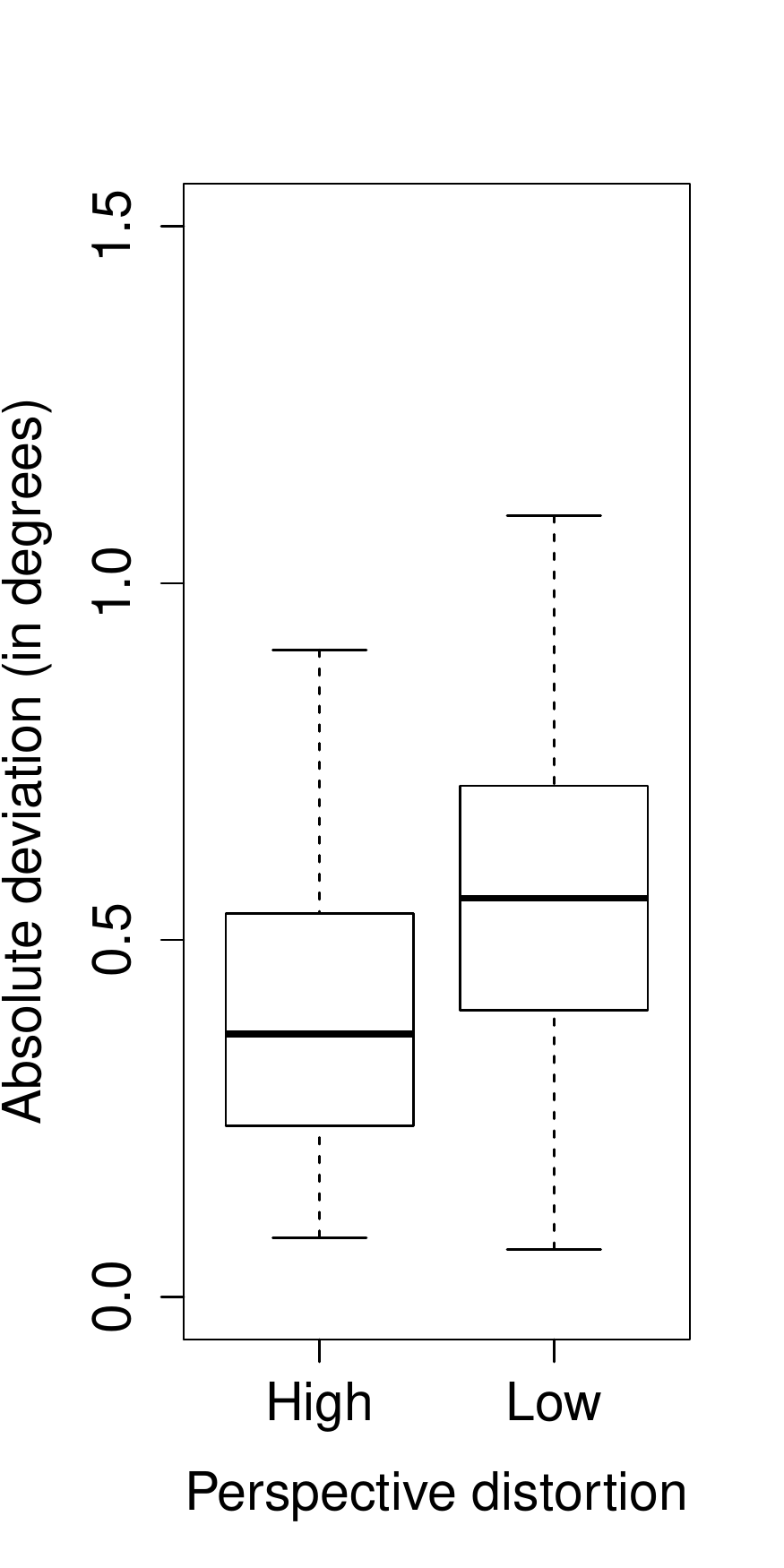}
  \end{minipage}
  \begin{minipage}{.245\linewidth}
    \includegraphics[width=\textwidth]{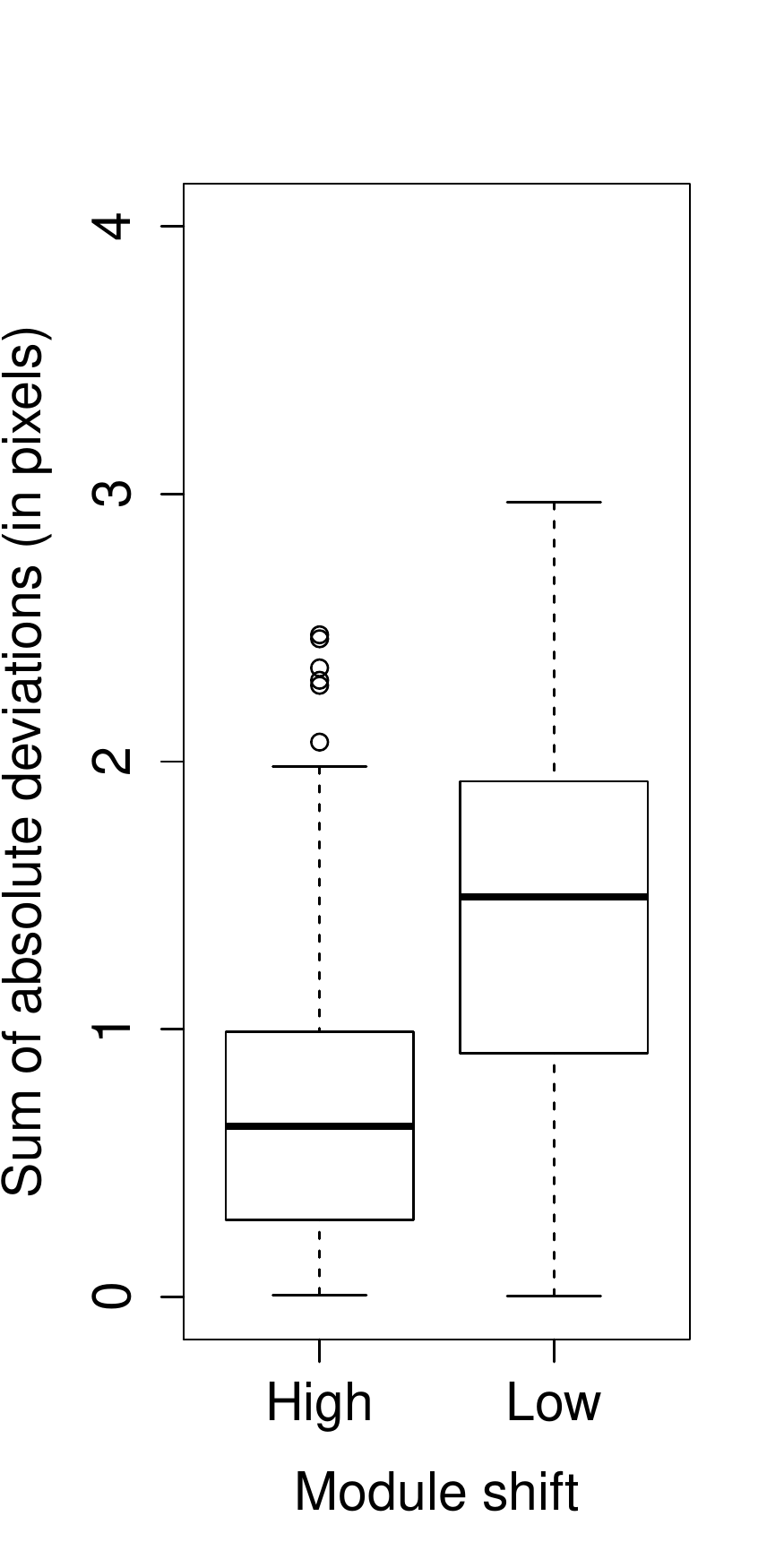}
  \end{minipage}
  \begin{minipage}{.245\linewidth}
    \includegraphics[width=\textwidth]{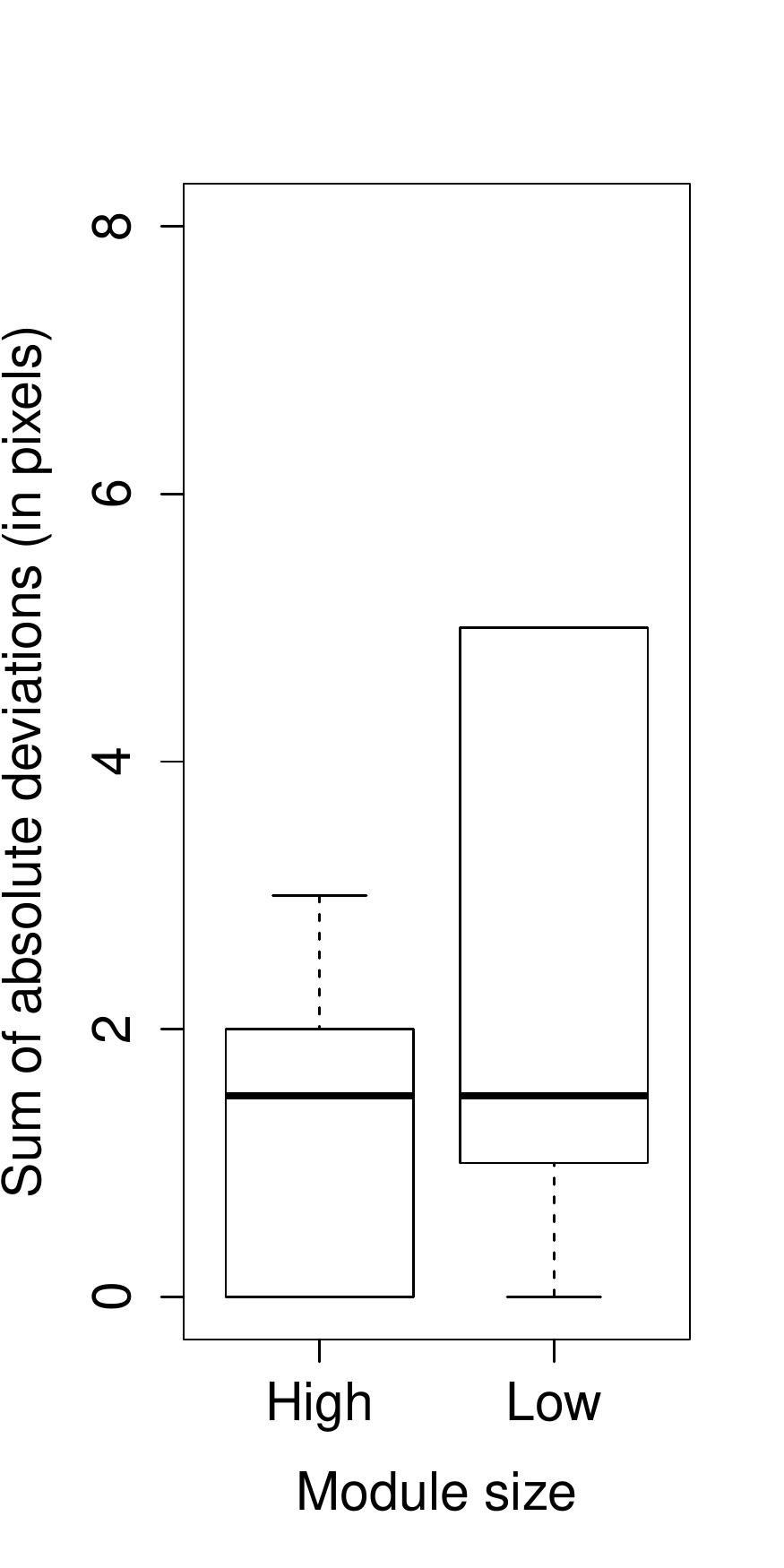}
  \end{minipage}
  \caption{Comparison of the performance for selected images with visually high and
    low contrast, respectively. Left to right: rotation distortion correction,
    perspective distortion correction and cell detection accuracy}
  \label{fig: high low contrast performance}
\end{figure*}

\section{Conclusions}
\label{sec:conclusions}

Motivated by the problem to preprocess and analyze electroluminescence
images of photovoltaic modules and detect all solar cells, this paper
proposes a comprehensive methodology for preprocessing image data by
correcting for distortions due to rotation, perspective, location and
size. Further, two methods are proposed to detect and extract cells
from an image, and the automatized algorithm outputs an image where
the extracted and standardized cells are put together. The methods
are optimized and to some extent tailor-made for photovoltaic
images. But they can be certainly adapted to similar industrial
imaging applications where the same problems arise. For example images of wafers consisting of electrical circuits
such as CPUs, which are also arranged at fixed distances. 

The approach combines specialized Hough transform algorithms,
statistical tools such as robust regression and change-point
estimation, as well as a priori knowledge from technical
specifications, in order to provide a reliable and automatized
work-flow to process such image data with high accuracy.

Extensive simulations were conducted to assess the accuracy of the
methods for a database of real images from photovoltaics. One may conclude from the numerical results
of these simulations that the overall procedure is accurate and reliably works even for massive
image data sets. There are only a few cases where the distortions and/or the module position and size can not
be recovered accurately. From a second simulation study, conducted for selected
low and high contrast images, we can conclude that the approach is capable of
dealing with low image quality as well. As a consequence, the approach reliably works for outdoor images and, when
mainstreaming the proposal in industrial production, one can expect
that it works well and reliable even under high-throughput conditions where ideal imaging conditions are difficult to guarantee. 

Future work could study the adaption to other solar cell technologies, especially thin-film solar cells, take into account imaging schemes where each EL image consists of several pictures, as well as extend and adopt the approach to similar industrial imaging problems.

\begin{ack}
  The authors would like to express their sincere appreciation for a
  grant from the German Ministry for Economic Affairs and Energie
  (BMWi), collaborative project PV-Scan, grant no. 0325588B. They
  thank Sunnyside upP, Cologne, International Solar Energy Research Center
  (ISC) Konstanz and T\"UV Rheinland Energie GmbH, Cologne, for
  providing the EL image data and discussing results.
  
  The comments and suggestions of anonymous reviewers are appreciated. 
\end{ack}

\begin{appendix}
\section{Appendix}
\label{sec:impl-deta}

Denote $d_{l} \coloneqq \sqrt{(1-\delta_{l})^{2} + \delta_{l}^{2}}$,
then conditions for the indicator function in (\ref{eq:3}) being
non-zero are given by
\begin{gather*}
  B_{R}((a,b)) \cap \mathcal{P}(x,y) \neq \emptyset
  \Leftrightarrow
  \exists l= 0,\ldots,(n-1): \\
    \left|
    a (1-\delta_{l}) + b \delta_{l} - x
  \right| \leq R d_{l},
\end{gather*}
subject to conditions
\[
  0 \leq a,b \leq W, L \leq b-a \leq U.
\]

\begin{gather*}
  \left\{
    \begin{array}{lll}
      b &\leq &\min\left(
                \frac{1 + Rd_{l}}{\delta_{l}}
                , W
                \right)\\
      b &\geq &\max\left(
                \frac{1 - W(1-\delta_{l}) - R d_{l}}{\delta_{l}}
                ,L
                \right)\\
      a &\leq &\min\left(
                \frac{1 - b \delta_{l} + R d_{l}}{1-\delta_{l}}
                , W, b-L
                \right)\\
      a &\geq &\max\left(
                \frac{1 - b\delta_{l} - R d_{l}}{1-\delta_{l}}
                , 0, b-U
                \right)
    \end{array}
  \right..
\end{gather*}
\end{appendix}




\bibliographystyle{ios1}
\bibliography{bibliography}







\end{document}